# FSS-Net: Frequency-Spatial Synergy Network with Wavelet Attention for Carotid Artery Ultrasound Segmentation

Jiawei Liu[1,2], Zhijiang Wan[2], Junhua Hu[3], Rongli Zhang[4], Zhongbiao Xu[5], Yankun Cao[6], Yuan Chen[7,*], Jin Hong[2,*]

1. Ji luan Academy, Nanchang University, Nanchang 330031, China
2. School of Information Engineering, Nanchang University, Nanchang 330031, China
3. State Key Laboratory of Water Cycle and Water Security, China Institute of Water Resources and Hydropower Research, Beijing 100038, China
4. Department of Diagnostic Radiology, Li Ka Shing Faculty of Medicine, The University of Hong Kong, Pok Fu Lam, Hong Kong, China
5. Department of Radiotherapy, Guangdong Provincial People's Hospital, Guangdong Academy of Medical Sciences, Southern Medical University, Guangzhou 510080, China
6. Joint SDU-NTU Centre for Artificial Intelligence Research (C-FAIR), Shandong University, Jinan 250101, China
7. Department of Pediatrics, Shandong Provincial Hospital Affiliated to Shandong First Medical University, Jinan 250021, China

E-mail: 8008123361@email.ncu.edu.cn; zhijiangwan@ncu.edu.cn; hujh@iwhr.com; rlzhang@hku.hk; xzbiao19890507@126.com; kunkun@sdu.edu.cn; yuanyuanchen01@126.com; hongjin@ncu.edu.cn;
* Correspondence should be addressed to Yuan Chen & Jin Hong

**Abstract:** Accurate segmentation of carotid arteries in ultrasound imaging is critical for stroke risk assessment. However, speckle noise, low contrast, and blurred boundaries remain major challenges. In this paper, we propose a Frequency-Spatial Synergy Network (FSS-Net) to achieve noise-robust and high-precision carotid artery segmentation. The network integrates wavelet transform, multi-domain attention, and edge enhancement into a unified encoder-decoder architecture. Specifically, a Channel-Spatial-Wavelet Attention (CSWA) module is designed to suppress noise and purify semantic features in the frequency domain. A Wavelet-Enhanced Bottleneck (WEB) module is introduced to capture long-range global dependencies efficiently. Furthermore, a Laplacian-Guided Adaptive Edge Fusion (LAEF) module compensates high-frequency details and maintains boundary continuity. Extensive experiments on carotid ultrasound datasets show that FSS-Net achieves a Dice score (DSC) of 96.46% and strong robustness under low SNR conditions, outperforming several state-of-the-art methods. This method realizes accurate segmentation of carotid artery in ultrasonic imaging, effectively identifies carotid atherosclerotic plaque, and is verified by other task (such as segmentation of breast cancer), suggesting that it has good clinical application potential in identifying abnormal tissue masses in ultrasonic images.

## 1. Introduction

Cardiovascular disease and stroke remain leading global causes of mortality, with carotid atherosclerotic plaques triggering approximately 18–25% of ischemic strokes [1-3]. Accurate measurement of these plaques is critical for risk assessment. While ultrasound is widely used for carotid screening [4-6], manual measurement and delineation are labor-intensive, driving the demand for automated measurement solutions to support emerging technologies like robotic ultrasound systems [7].

Deep learning has been widely applied in medical image analysis [8-10]. However, predominantly relying on spatial convolutions creates a bottleneck in low Signal-to-Noise Ratio (SNR) ultrasound measurement tasks, where speckle noise and vessel boundaries exhibit significant spatial overlap. This spatial entanglement forces an inherent trade-off: aggressive filtering blurs anatomical edges, compromising measurement precision, while preserving details inadvertently retains noise artifacts.

Despite these advancements, transverse carotid artery segmentation faces three persistent bottlenecks:

(i) **Noise Interference:** Inherent speckle noise and low contrast (**Fig. 1**) often obscure vessel boundaries. While some existing methods [11] employ wavelet transforms for preprocessing, they lack a deep, architectural integration of frequency analysis within the learning network.

(ii) **Boundary Dilution:** The gradual gray-scale transition between the lumen and vessel wall complicates edge definition. Continuous spatial downsampling in standard encoder-decoder architectures inevitably leads to the dilution of high-frequency edge information, compromising boundary continuity.

(iii) **Pathological Heterogeneity:** Pathologies such as calcified plaques (**Fig. 1**) disrupt normal anatomy with heterogeneous textures. Single-scale receptive fields often fail to reconcile global vessel morphology with local lesion details, resulting in segmentation uncertainty in abnormal regions.

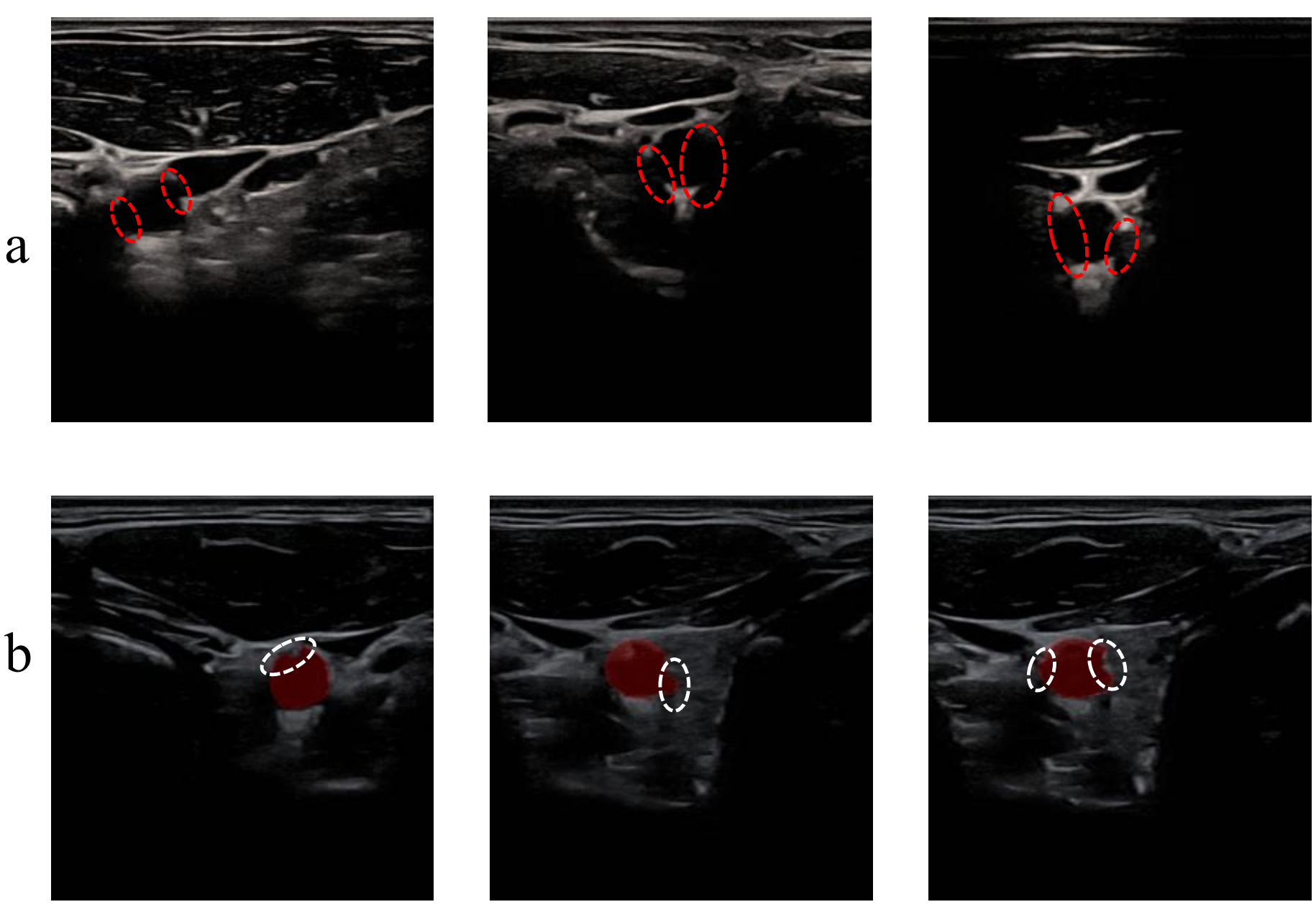

**Fig. 1** Carotid ultrasound images. (a) shows the fuzzy and discontinuous nature of the boundaries, and (b) shows abnormal anatomical structures.

Motivated by previous studies that combine wavelet transform with attention mechanisms [12-14], we further explore how frequency-domain representation can be integrated into a carotid ultrasound segmentation network. Rather than claiming that DWT alone can completely separate noise from anatomical structures, we use it as an explicit feature decomposition tool to represent low- and high-frequency components separately. This representation provides a useful basis for modeling the different roles of coarse vascular structure and boundary-related details in noisy ultrasound images. Based on this motivation, we construct a frequency-spatial collaborative framework consisting of three complementary modules:

First, the Channel-Spatial-Wavelet Attention (CSWA) module is introduced in the encoder to perform joint channel, spatial, and frequency-aware feature modulation. Instead of relying solely on spatial responses, CSWA uses wavelet decomposition to represent low-frequency structural information and high-frequency detail information separately. The low-frequency sub-band provides a relatively stable structural cue for attention generation, while the high-frequency sub-bands are retained during reconstruction to preserve boundary-related details and reduce the influence of scattered noise-like responses. Building on this purified representation, the Wavelet-Enhanced Bottleneck (WEB) module operates as a Global Context-Aware unit at the bottleneck. It utilizes multi-pooling strategies and large-kernel convolutions within the wavelet domain to capture long-range dependencies, ensuring robustness against heterogeneous plaque textures. Finally, the Laplacian-Guided Adaptive Edge Fusion (LAEF) module is introduced in the decoder to refine boundary-related responses during feature reconstruction. Although high-frequency sub-bands are retained in the CSWA reconstruction, fine edge cues may still be weakened by repeated downsampling and nonlinear feature transformations. LAEF therefore incorporates Laplacian edge priors into the skip fusion process and adaptively combines them with semantic decoder features, aiming to improve boundary localization and contour continuity. The main contributions of this work are as follows:

(i) *Frequency-Spatial Synergistic Architecture*: We present FSS-Net, an encoder-decoder framework for carotid artery ultrasound segmentation. The framework incorporates wavelet-based frequency-domain representation into spatial feature learning, allowing the network to model structural cues and detail-sensitive responses in a complementary manner. This design is intended to improve segmentation stability in ultrasound images with speckle noise, low contrast, and indistinct vessel boundaries.

(ii) *Triple-Domain Attention Mechanism*: The proposed CSWA module integrates channel, spatial, and frequency-domain attention, achieving multiscale feature disentanglement and cross-domain contextual mining. Unlike conventional attention modules (e.g., BAM, CBAM), it uses wavelet decomposition to represent structural and detail-related components separately, improving feature discrimination in low-contrast scenarios.

(iii) *Adaptive Edge-Semantic Fusion*: The LAEF module addresses the challenge of blurred boundaries by explicitly extracting edge features via Laplacian kernels and dynamically fusing them with deep semantic features through an adaptive weighting mechanism. This ensures the model prioritizes boundary continuity without sacrificing high-level contextual understanding, outperforming classic edge enhancement strategies (e.g., Attention Gate, Res-Block).

(iv) *Global Context-Aware Bottleneck*: The WEB module optimizes the network's bottleneck layer by integrating wavelet decomposition, multi-pooling strategies, and large-kernel depth-wise convolutions. It captures global contextual information across statistical and frequency dimensions, enhancing the model's robustness to heterogeneous plaque textures (e.g., calcifications) and long-range spatial dependencies.

(v) *Clinically Validated Superiority*: FSS-Net achieves competitive performance (DSC = 96.46%) on the Common Carotid Artery Ultrasound Dataset, with exceptional noise robustness (only 0.2% DSC drop under 19.28 dB SNR) and cross-task generalization (breast ultrasound segmentation DSC= 75.86%). Its lightweight design (5.01M parameters) facilitates clinical deployment, providing a reliable tool for automated plaque measurement and stroke risk assessment.

## 2. Related work

### 2.1. Segmentation of transverse carotid-artery images

Early segmentation approaches predominantly relied on active contours and morphological operators [15-17] to delineate arterial boundaries. While effective in high-contrast scenarios, these traditional methods lack robustness against the intrinsic speckle noise and artifacts of ultrasound imaging. The advent of deep learning has shifted the paradigm towards Convolutional Neural Networks (CNNs). Recent works have employed Faster R-CNN for ROI localization [18] or specialized architectures like DoubleUNet-BAM [19] and deeply supervised FCNs [20] for stenosis diagnosis. However, these deep learning methods typically operate exclusively within the spatial domain. A fundamental limitation arises from this reliance: spatial convolutions struggle to disentangle multiplicative speckle noise from tissue texture due to their overlapping intensity distributions. Consequently, these models often face an inherent trade-off—aggressive smoothing to remove noise sacrifices edge fidelity, while preserving details inevitably retains noise artifacts. Furthermore, most existing architectures lack explicit, end-to-end mechanisms for preserving fine-grained boundary topology, leading to performance degradation in low-SNR clinical environments.

### 2.2. Edge detection

Traditional edge detection mainly relies on low-level image cues (gradients, textures, color discontinuities), with classical operators (Sobel, Canny, Laplacian) localizing edges via first- or second-order intensity derivatives. Though computationally efficient, they are noise-sensitive and fail to capture semantic boundaries in complex scenes. To address these issues, recent studies integrate edge detection with deep learning: Wang et al. [21] proposed the EAP module for noise suppression and edge cue extraction, refined via MPR to reduce boundary blurring. Similarly, recent advancements in instrumentation have introduced edge feature-induced networks for cardiac segmentation and attentive boundary-aware transformers for defect detection. These studies collectively demonstrate that explicitly enforcing boundary constraints is crucial for high-precision segmentation in low-contrast imagery [22, 23]; Sun et al. [24] adopted a pyramid architecture enhanced by the CSE module for multi-scale edge extraction, boosted by a multi-path down-sampling block.

While explicit edge constraints are indispensable for accurate segmentation, traditional descriptors are inherently compromised by ultrasound speckle noise. Moreover, existing deep learning integrations often suffer from suboptimal fusion strategies, where edge detection is relegated to an auxiliary task, resulting in the progressive dilution of boundary semantics during decoding. To transcend these limitations, we propose the Laplacian-Guided Adaptive Edge Fusion (LAEF) module. Unlike static or naive fusion paradigms, LAEF establishes a dynamic pathway that leverages high-level semantic feedback to suppress spurious noise gradients. By integrating explicit edge priors via an adaptive weighting mechanism, it significantly enhances the discriminability of ambiguous boundaries and enforces topological continuity, all while maintaining high computational efficiency.

### 2.3. Wavelet attention

Attention mechanisms are a cornerstone for enhancing CNNs' representational power by adaptively highlighting informative and suppressing redundant features. Park et al. [25] proposed the BAM, decoupling attention inference into parallel channel and spatial branches; Woo et al. [26] developed the CBAM to sequentially apply channel and spatial attention for refined feature extraction. Subsequently, researchers integrated wavelet transforms with attention mechanisms: Zhao et al. [27] proposed the WA module to perform attention on high-frequency sub-bands while preserving low-frequency structure; Cheng et al. [28] and Qi et al. [29] combined them with spatial attention for superior retinopathy detection; Tian et al. [30] integrated residual-in-residual dense blocks with dual attention on high-frequency sub-bands for ultrasound denoising, increasing Peak Signal-to-Noise Ratio (PSNR) by 2.3 dB; Xu et al. [31] used Haar-wavelet downsampling instead of max-pooling with embedded channel attention, improving Cityscapes Intersection over Union (IoU) by 1.8%. These studies confirm that combining wavelet transforms and attention mechanisms enhances both detail preservation and semantic consistency.

Existing wavelet-based attention and segmentation methods have explored frequency-domain representation from different perspectives. Wave-ViT [32] integrates wavelet-based invertible downsampling into Transformer attention to reduce information loss caused by aggressive key/value downsampling and improve the efficiency–accuracy trade-off in general visual recognition tasks. HWD [33] replaces conventional pooling or strided convolution with Haar wavelet downsampling, aiming to preserve more spatial information during encoder-side downsampling. In medical image segmentation, WRANet [34] introduces DWT into a U-Net-like architecture by replacing standard downsampling modules and removing high-frequency components to reduce noise interference. MISegNet [35] further incorporates DWT-based frequency-domain features and a self-attention-based global context-aware module to enlarge the receptive field for multi-modal medical image segmentation. These studies demonstrate the potential of wavelet-domain modeling for feature preservation, denoising, and contextual representation.

However, most existing wavelet-based methods mainly focus on efficient downsampling, high-frequency enhancement/suppression, or frequency-domain refinement at a specific network stage. In contrast, FSS-Net introduces a coordinated frequency-aware design in both the encoder and bottleneck. In the encoder, CSWA uses the LL sub-band as a relatively stable structural cue for attention generation, while retaining LH, HL, and HH sub-bands during reconstruction. This differs from methods that simply discard high-frequency components, because the high-frequency sub-bands may contain both noise-sensitive fluctuations and boundary-related directional details. At the bottleneck, WEB further performs wavelet-domain contextual aggregation by decomposing high-level features into multiple sub-bands and applying multi-pooling-based spatial attention with large-kernel convolution. This design aims to enhance global contextual representation while preserving frequency-specific information at the deepest semantic layer. Therefore, compared with existing wavelet-attention or wavelet-downsampling methods, the main distinction of FSS-Net lies in the combination of LL-guided encoder attention and wavelet-enhanced bottleneck context modeling for low-contrast carotid ultrasound segmentation.

## 3. Method

### 3.1. Overall architecture

The overall architecture of FSS-Net is illustrated in **Fig. 2**. Built upon the classical U-Net [36] framework, the network comprises three pivotal components: the Channel-Spatial-Wavelet Attention (CSWA), Laplacian-Guided Adaptive Edge Fusion (LAEF) module, and Wavelet-Enhanced Bottleneck (WEB).

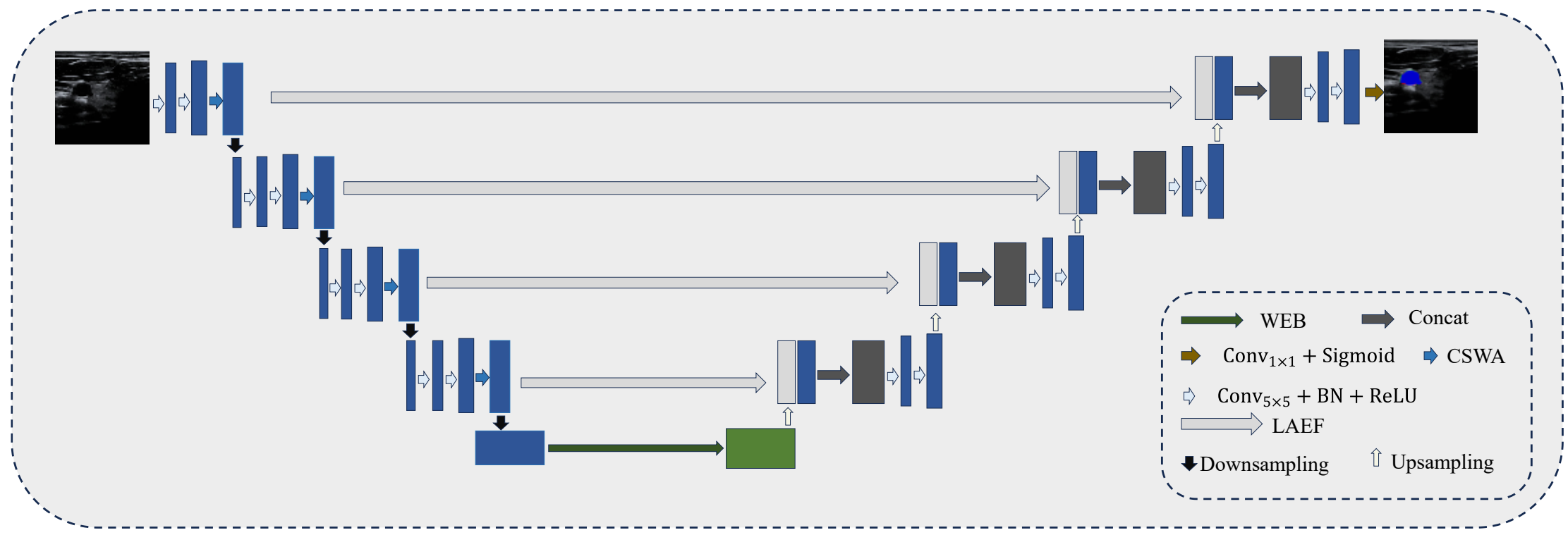

**Fig. 2** The overall framework of FSS-Net

The encoding path begins with a 256×256 input image processed through four cascaded Wavelet-encoder stages. Structurally, each stage follows a sequential design: a ConvBlock (comprising two 5×5 convolutions with Batch Normalization and ReLU) first extracts local features, followed immediately by the CSWA module, and concluding with a downsampling operation. Crucially, the CSWA module is strategically positioned before downsampling to explicitly disentangle noise artifacts from semantic features in the frequency domain while high spatial resolution is still preserved.

Subsequently, the 16×16, 256-dimensional features output by the encoder are fed into the WEB. This module refines the features in the wavelet domain: the channel count is first expanded to 1024, the tensor is then decomposed by Discrete Wavelet Transform (DWT) into multiple frequency sub-bands, each of spatial size 8×8 and 1024 channels. A spatial-attention block is applied to selectively enhance every sub-band, highlighting the most informative frequency components. Finally, the enhanced sub-bands are reconstructed via Inverse Wavelet Transform (IWT), and a 1×1 convolution compresses the channels back to 256, yielding a polished high-level feature map for the decoder.

Finally, the refined 16 × 16, 256-dimensional feature volume is fed into the decoder. At each decoder stage, a transposed convolution first up-samples the activation map, doubling its spatial resolution while halving its channel dimensionality. To counterbalance potential high-frequency information loss during encoding, we replace conventional skip connections with the LAEF module. LAEF explicitly extracts edge priors via Laplacian kernels and adaptively fuses them with up-sampled features, ensuring topological continuity. A subsequent ConvBlock integrates and refines the combined cues, compressing the channels to the desired dimension for the next decoding stage. Finally, the decoder outputs a 256×256, 16-dimensional feature map, which, after being compressed into a single channel via 1×1 convolution, generates a segmentation probability map consistent with the input size through the Sigmoid activation function.

### 3.2. Channel-Spatial-Wavelet Attention (CSWA)

Carotid ultrasound images are often affected by speckle noise, low contrast, and blurred vessel boundaries, which may limit the reliability of purely spatial attention mechanisms. Since spatial attention is mainly derived from local feature responses, noise-like intensity fluctuations may be incorrectly emphasized together with true boundary cues. To alleviate this problem, we introduce the CSWA module as a frequency-aware feature modulation mechanism. Specifically, DWT is used as an interpretable decomposition operation rather than a hard noise-separation process. The LL sub-band mainly provides stable structural information, whereas LH, HL, and HH capture direction-sensitive high-frequency responses that include both boundary-related details and noise-sensitive local fluctuations. Based on this representation, CSWA uses low-frequency structural cues to guide attention generation while retaining high-frequency sub-bands during reconstruction.

The overall framework of the proposed CSWA module is illustrated in **Fig. 3**. Specifically, we first decompose the features processed by the ConvBlock into one low-frequency sub-band (LL) and three high-frequency sub-bands (LH, HL, HH) using the Daubechies-4 (db4) wavelet transform. The LL sub-band encodes global structural information, while the LH, HL, and HH sub-bands capture horizontal, vertical, and diagonal detail features, respectively. Mathematically, the discrete wavelet transform (DWT) employed herein can be formulated as follows:

$$W_{j,k} = \iint f(x,y)\varphi_{j,k}(x,y)dxdy \tag{1}$$

here $f(x,y)$ is the input image, $\varphi_{j,k}(x,y)$ is the wavelet basis function, and $W_{j,k}$ are the wavelet coefficients representing the features at specific scales and positions. Similarly, the original image can be reconstructed using the Inverse Discrete Wavelet Transform (IDWT):

$$f(x,y) = \sum_{j,k} W_{j,k}\varphi_{j,k}(x,y) \tag{2}$$

For the low-frequency sub-band (LL), its feature map is denoted as $F \in R^{H\times W\times C}$. To generate channel attention weights, we first perform global average pooling (GAP) and global max pooling (GMP) on $F$, followed by processing via a shared multi-layer perceptron (MLP). Specifically, the shared MLP implements channel compression and expansion through 1×1 convolutions. Finally, the two pooled results are fused using a Sigmoid activation function to generate channel attention weights, which are applied for channel-wise weighting of the input feature map. This operation suppresses irrelevant channels and highlights critical channel information. The corresponding mathematical formulations are as follows:

$$MLP(x) = Conv_{1\times1}(ReLU\left(Conv_{1\times1}\left(x, C \to \frac{C}{r}\right)\right), \frac{C}{r} \to C) \tag{3}$$

$$F_1 = F \odot \sigma(MLP(AvgPool(F)) + MLP(MaxPool(F))) \tag{4}$$

where $F_1$ denotes the channel-wise weighted feature map, $Conv_{1\times1}$ represents a 1×1 convolution, $\odot$ denotes element-wise multiplication, $C \to \frac{C}{r}$ denotes reducing the number of channels from $C$ to $\frac{C}{r}$, $\sigma$ denotes the Sigmoid activation function, and $ReLU$ denotes a nonlinear activation function.

Subsequently, we feed the channel-wise weighted feature map $F_1 \in R^{H\times W\times C}$ into the incremental wavelet convolution module. Specifically, in this module, we convolve the LL sub-band obtained from each decomposition. From the second decomposition onwards, only the LL sub-band derived from the previous level is further decomposed, and the same applies to the third decomposition. After each wavelet convolution operation, a Batch Normalization (BN) layer is added to balance the contribution of each branch.

Thereafter, the outputs of different branches are summed, followed by channel shuffle to facilitate inter-channel communication. We then reduce the dimension of the fused feature map to 1 via a $Conv_{1\times1}$ layer and generate the spatial attention map $H(F) \in R^{H\times W\times 1}$ using the Sigmoid activation function. Subsequently, the spatial attention map is duplicated along the channel dimension to obtain $H'(F) \in R^{H\times W\times C}$, and finally element-wise multiplied with the input feature $F_1$. In summary, the calculation of our multi-scale channel-wavelet spatial attention is formulated as follows:

$$H(F) = \sigma(W_{cov}(\sum_{i=1}^{n} BN\ (W_T^n(F_1)))) \tag{5}$$

$$F_2 = F + F \odot H'(F) \tag{6}$$

where $F_2$ denotes the final output of the LL sub-band, $H'(F)$ denotes the spatial attention map, $W_{cov}$ denotes the 1×1 conventional convolution, and $W_T^n$ denotes the n-level wavelet

convolution—where the value of n depends on the size of the feature map.

Finally, to avoid the loss of details, we sum the LL sub-band processed by the multi-scale channel-wavelet spatial attention with the other three sub-bands, and perform inverse wavelet transform (IWT) to restore the features in the spatial domain. The corresponding mathematical formulation is as follows:

$$F_{out} = IWT(LL' + LH + HL + HH) \tag{7}$$

where $F_{out}$ denotes the final output feature of the CSWA module, and $LL'$ denotes the processed low-frequency sub-band.

**Algorithm 1** summarizes the complete computational pipeline of the proposed CSWA module.

**Algorithm 1** CSWA Module

Input: Feature map $F \in R^{H \times W \times C}$

Output: Enhanced feature map $F_{out} \in R^{H \times W \times C}$

1: Wavelet Decomposition:

Perform Daubechies-4 DWT on F:

$LL, LH, HL, HH = DWT(F)$

2: Channel Attention on LL Sub-band:

$F_{att} = LL \odot Sigmoid(SharedMLP(GAP(LL)) + SharedMLP(GMP(LL)))$

3: Multi-scale Wavelet Recursive Convolution:

For each scale level $l \in \{1,2,3\}$:

Apply wavelet decomposition and recursive convolution to $LL$

End

4: Sum outputs from all scales, normalize

5: Channel Shuffle to promote cross-channel interaction

6: Generate Spatial Attention Map:

$S_{att} = \sigma(Conv_{1\times1}(aggregated\ features))$

7: Weighted Fusion:

$LL_{enhanced} = F_{att} + F_{att} \odot S_{att}$

8: Reconstruction:

$F_{out} = IDWT(LL_{enhanced} + LH + HL + HH)$

Return $F_{out}$

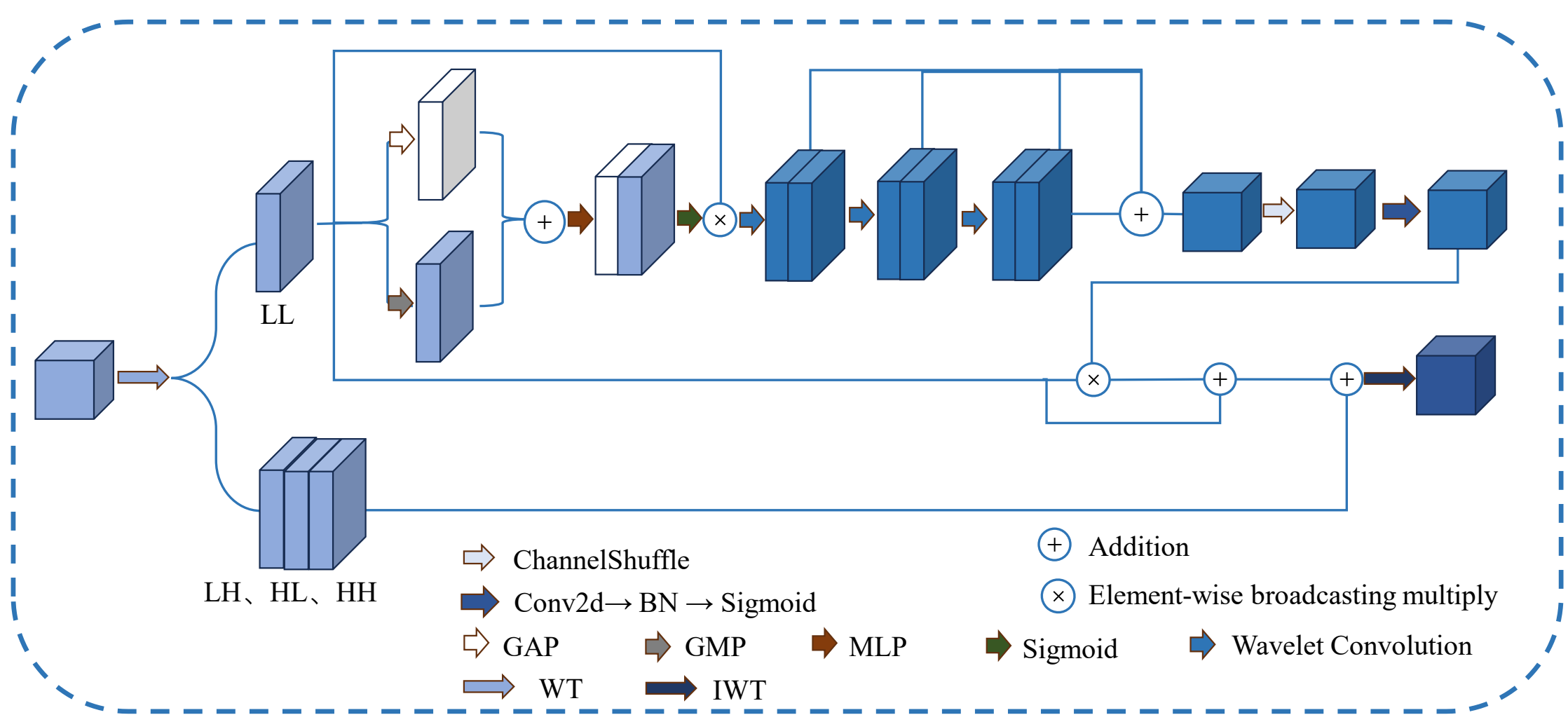

**Fig. 3** The overall framework of CSWA

### 3.3. Laplacian-Guided Adaptive Edge Fusion (LAEF)

Although CSWA retains high-frequency sub-bands during wavelet reconstruction, fine boundary responses may still be weakened by repeated downsampling and nonlinear feature transformations in the encoder-decoder architecture. This issue is particularly relevant for carotid ultrasound images, where vessel-wall boundaries are often blurred and low-contrast. Therefore, we designed the LAEF module (**Fig. 4**) to supplement the decoder with anatomical boundary responses. Since raw edge features are shallow and may interfere with deep semantic learning if directly fused, we adopt an adaptive fusion mechanism inspired by SK-Net [37], which dynamically fuses explicit edge priors with encoder features to retain boundary topology and deep semantic representations, thereby improving segmentation performance. The specific steps are as follows:

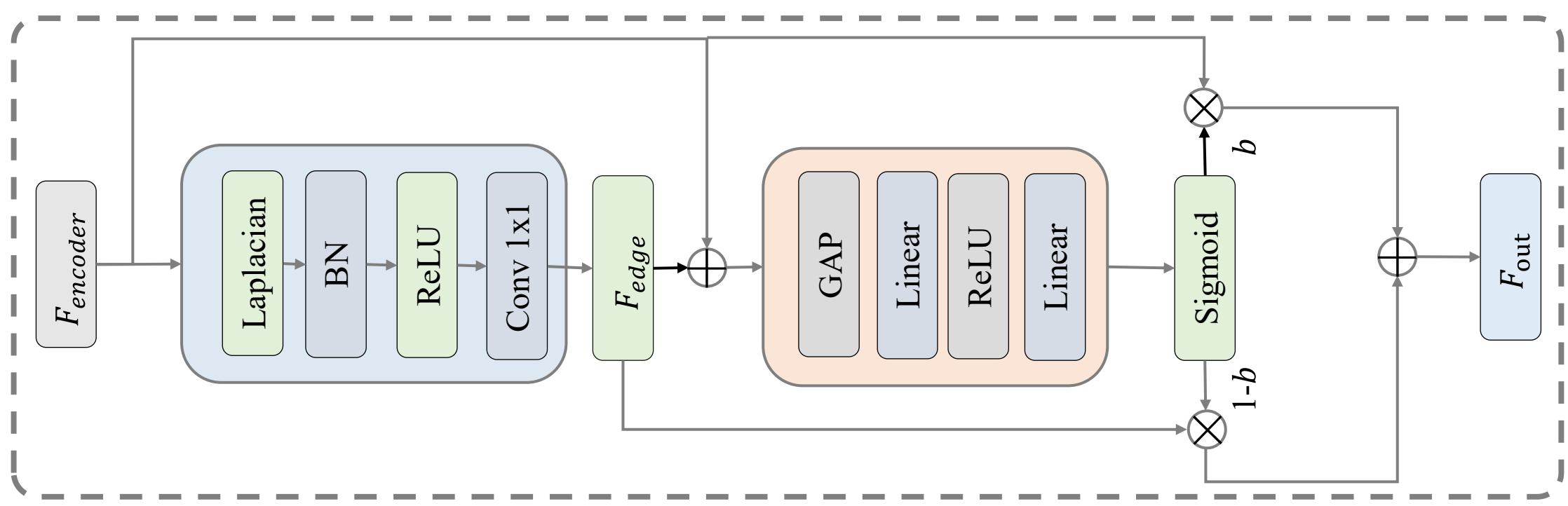

**Fig. 4** The overall framework of LAEF

First, we perform edge extraction using the Laplacian operator. As the responses of the Laplacian operator exhibit rotation invariance, it is suitable for the circular-like shape of carotid artery cross-sections. For the Laplacian kernel, we adopt the classic 3×3 edge detection kernel:

$$\begin{bmatrix} -1 & -1 & -1 \\ -1 & 8 & -1 \\ -1 & -1 & -1 \end{bmatrix}$$

Next, we apply a normalization layer and an activation layer to the extracted raw edge features to introduce non-linearity. Then, we adjust the channel dimension through a 1×1 convolution to ensure compatibility between the edge features and the input of subsequent modules, and finally output the edge feature $F_{edge}$. Subsequently, we perform element-wise addition fusion of $F_{encoder}$ and $F_{edge}$ to obtain $F_{fused}$, which serves as the basis for calculating attention weights. The calculation formula is as follows:

$$F_{fused} = F_{edge} + F_{encoder} \tag{8}$$

Then, we perform spatial dimension compression on $F_{fused}$ through Global Average Pooling (GAP) to obtain channel-wise statistical features. Then, we generate channel attention weights $b \in [0, 1]$ via two fully connected layers with hidden layer dimensions of $\frac{C}{2}$ and a Sigmoid activation function. Finally, we achieve weighted fusion of features based on the attention weights:

$$F_{out} = F_{encoder} \cdot b + F_{edge} \cdot (1 - b) \tag{9}$$

where $F_{out}$ denotes the final output feature.

**Algorithm 2** summarizes the complete computational pipeline of the proposed LAEF module.

---

**Algorithm 2** LAEF Module

---

Input:

Encoder feature $F_{enc} \in R^{H \times W \times C}$

Decoder feature $F_{dec} \in R^{H \times W \times C}$

Output:

Fused feature $F_{out} \in R^{H \times W \times C}$

Edge feature $F_{edge} \in R^{H \times W \times C}$

1: Laplacian Edge Extraction:

Apply depthwise convolution with Laplacian kernel

$$K = \begin{bmatrix} -1 & -1 & -1 \\ -1 & 8 & -1 \\ -1 & -1 & -1 \end{bmatrix}$$

$$F_{edge} = DepthwiseConv(F_{enc}, K)$$

2: Normalization & Activation:

$$F_{edge} = ReLU(BatchNorm(F_{edge}))$$

3: Feature Fusion for Attention Calculation:

$$F_{fused} = F_{enc} + F_{edge}$$

4: Channel-wise Statistics:

$$g = GlobalAvgPool(F_{fused})$$

5: Adaptive Channel Attention:

$$b = \sigma(W_2 \cdot ReLU(W_1 \cdot g))$$

6: Weighted Fusion:

$$F_{out} = b \odot F_{att} + (1 - b) \odot F_{edge}$$

Return $F_{out}, F_{edge}$

---

**3.4. Wavelet-Enhanced Bottleneck (WEB)**

In the encoder-decoder architecture, the bottleneck layer represents the zenith of semantic abstraction, acting as the core hub for feature transitions. However, standard implementations typically suffer from low spatial resolution, and traditional convolutional layers with fixed, local receptive fields struggle to capture the global morphology of the vessel cross-section. This limitation is particularly critical when attempting to establish segmentation continuity across heterogeneous pathologies, such as calcified plaques with acoustic shadows. To address these challenges, we propose the WEB module (**Fig. 5**). By shifting the processing paradigm to the frequency domain via the Discrete Wavelet Transform (DWT), this module serves a triple purpose: it efficiently compresses feature dimensions, actively purifies effective semantic information from inherent speckle noise, and enhances the global representation of core vascular structures. This design ensures that the network retains a robust global context even when local anatomical cues are disrupted by artifacts.

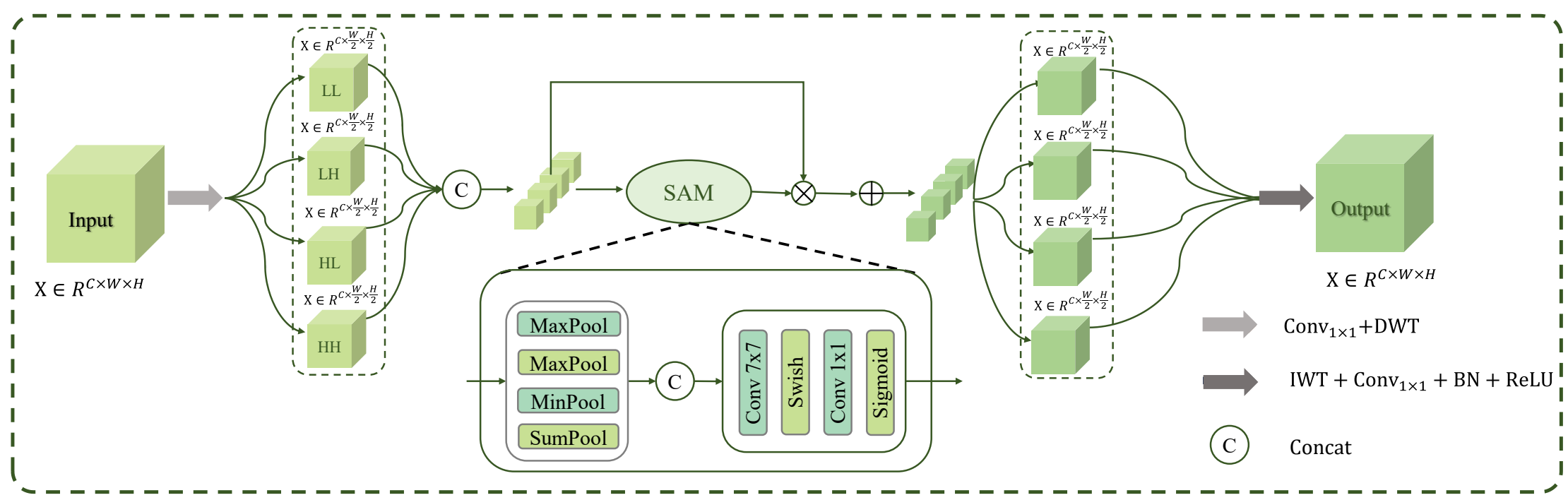

**Fig. 5** The overall framework of WEB

First, we perform channel expansion on the high-dimensional feature map $X \in R^{16\times16\times C}$ (the final output of the encoder) via a 1×1 convolution to obtain the feature $X_2 \in R^{16\times16\times4C}$. This ensures that sub-band concatenation after subsequent wavelet decomposition does not cause an information bottleneck, while providing sufficient channel capacity for the attention mechanism. Then, we adopt the Daubechies-4 wavelet basis to perform DWT for frequency-domain decomposition of $X_2$, yielding the low-frequency sub-band (LL), horizontal high-frequency sub-band (LH), vertical high-frequency sub-band (HL), and diagonal high-frequency sub-band (HH). These four sub-bands are concatenated along the channel dimension to obtain $X_3 \in R^{8\times8\times16C}$. To further enhance the model's ability to capture global dependency of features, we introduce the Spatial Attention Module (SAM) [38] to explore long-range dependencies. Within SAM, we apply four spatial pooling operations (MeanPool, MaxPool, MinPool, SumPool) on $X_3$ to obtain $P_{mean}$, $P_{max}$, $P_{min}$, $P_{sum} \in R^{8\times8\times1}$, which are then concatenated into $P_{cat} \in R^{8\times8\times4}$. Next, a 7×7 depthwise convolution is used to capture long-range spatial dependencies, and the Swish activation function is incorporated to introduce nonlinearity. Subsequently, a 1×1 convolution compresses $P_{cat}$ into a single-channel spatial attention weight $W_{att} \in R^{8\times8\times1}$. The attention weight is expanded via a broadcasting mechanism to match the dimension of $X_3$, and then normalized to the interval [0, 1] using the Sigmoid activation function. Finally, residual weighted fusion is performed to ensure that the original sub-band information is not excessively suppressed, achieving key region feature enhancement. Meanwhile, a residual connection is added to avoid feature information loss. The corresponding formula is as follows:

$$X_{enh} = X_3 \odot W_{att} + X_3 \tag{10}$$

where $X_{enh}$ denotes the enhanced feature.

Subsequently, we split $X_{enh}$ into 4 enhanced sub-bands $LL_{enh}$, $LH_{enh}$, $HL_{enh}$, $HH_{enh}$ along the channel dimension. These sub-bands are then reconstructed via Inverse Wavelet Transform (IWT) into a feature map $X_{out}$ with the same size as the expanded feature. The corresponding mathematical expression is as follows:

$$Y = ReLU(BN(Conv_{1\times1}(X_{enh}))) \tag{11}$$

The complete computational workflow of the WEB module is formalized in **Algorithm 3**.

**Algorithm 3** WEB Module

Input: High-level feature $X \in R^{16\times16\times C}$

Output: Refined feature $Y \in R^{16\times16\times C}$

1: Channel Expansion:

$X_{exp} = Conv_{1\times1}(X)$ // Expand channels to 4C

2: Wavelet Decomposition:

Perform DWT on $X_{exp}$ using Daubechies-4 wavelet:

$LL, LH, HL, HH = DWT(X_{exp})$

3: Sub-band Concatenation:

$X_{exp} = Concat(LL, LH, HL, HH)$ // Shape: 8×8×16C

4: Spatial Attention Enhancement:

// Apply four spatial pooling operations

$P_{mean} = MeanPool(X_{sub})$

$P_{max} = MaxPool(X_{sub})$

$P_{min} = MinPool(X_{sub})$

$P_{sum} = SumPool(X_{sub})$

$P_{cat} = Concat(P_{mean}, P_{max}, P_{min}, P_{sum})$

5: Depthwise Convolution & Activation:

$P_{dw} = Swish(DepthwiseConv_{1\times1}(P_{cat}))$

6: Spatial Attention Weight Generation:

$W_{att} = \sigma(Conv_{1\times1}(P_{dw}))$

7: Attention-weighted Enhancement:

$X_{enh} = X_{sub} \odot W_{att} + X_{sub}$

8: Sub-band Splitting:

$Split(X_{enh}) \rightarrow LL_{enh} + LH_{enh} + HL_{enh} + HH_{enh}$

9: Wavelet Reconstruction:

$X_{rec} = IWT(LL_{enh}, LH_{enh}, HL_{enh}, HH_{enh})$

10: Channel Compression:

$Y = ReLU(BatchNorm(Conv_{1\times1}(X_{rec})))$

Return $Y$

### 3.5. Loss Function

The loss function combines Binary Cross Entropy Loss (BCE Loss), DSC Loss, and Boundary Loss: BCE loss provides stable gradient signals by comparing predicted probabilities with ground truth labels pixel by pixel. It is represented as:

$$Loss_{BCE} = -\frac{1}{N}\sum_{i=1}^{N}[y_i \log(p_i) + (1 - y_i)\log(1 - p_i)] \quad (12)$$

where $N$ denotes the batch size, $y_i$ is the true class label of the i-th sample and takes values of 0 or 1, and $p_i$ represents the model-predicted probability of sample *i*.

DSC loss optimizes the overlap of segmentation regions by calculating the overlap between predicted and actual segmentation areas. It is represented as:

$$Loss_{DSC} = 1 - \frac{2\times TP}{2\times TP+FP+FN} \quad (13)$$

Boundary loss measures boundary matching quality and enhances edge detection capability by computing the Dice similarity between the model's extracted edge features and the ground truth boundary masks.

$$Loss_{Boundary} = 1 - \frac{2\sum_{i=1}^{N} P_{edge,i} \cdot G_{edge,i}}{\sum_{i=1}^{N} P_{edge,i} + \sum_{i=1}^{N} G_{edge,i}} \quad (14)$$

where $P_{edge,i}$ and $G_{edge,i}$ denote the predicted edge probability and the ground truth boundary label at the $i$-th pixel, respectively. $N$ represents the total number of pixels in the image.

These three losses are equally weighted and summed to form the total loss function. The calculation formula of total loss is as follows:

$$Loss_{total} = Loss_{DSC} + Loss_{BCE} + Loss_{Bourndary} \tag{15}$$

## 4. Experiments and Results

### 4.1. Dataset

Common Carotid Artery Ultrasound Images Dataset [39]: The Common Carotid Artery Ultrasound Images Dataset was used as the primary benchmark in this study. This dataset is publicly available from Mendeley Data under a CC BY 4.0 license. It contains a total of 1,100 B-mode ultrasound images acquired from 11 subjects using a Mindray UMT-500Plus scanner equipped with an L13-3s linear probe. The original DICOM series were converted into PNG format and cropped to a resolution of 709×749 pixels. Each subject contributed 100 images, covering both left and right carotid arteries. Crucially, the dataset provides pixel-level ground truth masks, which were manually delineated by technicians and subsequently verified by clinical experts to ensure anatomical accuracy for segmentation tasks.

Breast Ultrasound Images Dataset [40]: The Breast Ultrasound Images Dataset, also known as BUSI, was used as an auxiliary dataset to evaluate the generalization ability of the proposed method on a different ultrasound segmentation task. It is publicly available and was originally collected at Baheya Hospital for Early Detection and Treatment of Women's Cancer, Cairo, Egypt. The dataset contains 780 breast ultrasound images from 600 female patients aged between 25 and 75 years, acquired in 2018 using LOGIQ E9 and LOGIQ E9 Agile ultrasound systems with ML6-15-D Matrix linear probe transducers. The images have an average size of approximately 500 × 500 pixels and are categorized into three classes: 133 normal, 437 benign, and 210 malignant images. Since normal cases do not contain lesion foregrounds, they were excluded from the segmentation experiments. Thus, only 647 lesion-annotated images, including 437 benign and 210 malignant cases, were used in this study. The corresponding ground-truth masks were provided with the dataset and were generated through manual annotation and expert review.

### 4.2. Implementation details

The proposed method was implemented using the PyTorch framework and executed on an NVIDIA RTX 4090 GPU. Prior to training, all input images were resized to 256×256 pixels and pre-processed using instance-level Z-score normalization (subtracting the mean and dividing by the standard deviation per image) to accelerate convergence. Both datasets were evaluated using five-fold cross-validation with a fixed random seed. For the Common Carotid Artery Ultrasound Images Dataset, all 1,100 images were used. In each fold, four folds were used for training and the remaining fold was used for testing. For the Breast Ultrasound Images Dataset, normal cases were excluded because they do not contain lesion foregrounds; only benign and malignant images with corresponding lesion masks were used for lesion segmentation. The same five-fold protocol was applied to this dataset. The network was optimized using the AdamW optimizer with an initial learning rate of 0.001 and a batch size of 16, with training extending for 50 epochs per fold. During each fold, the model achieving the best DSC on the held-out fold was selected and saved. Consequently, the final results represent the average of the 5-fold tests, with the standard deviation calculated to provide robust performance metrics. To quantitatively evaluate the effectiveness of the model, we employed the Dice Similarity Coefficient (DSC), Intersection over Union (IoU), Precision, Recall, Accuracy, Specificity (Spec), Matthews Correlation Coefficient (MCC) and the 95th percentile Hausdorff Distance (HD95). These metrics are derived based on True Positives (TP), True Negatives (TN), False Positives (FP), and False Negatives (FN). In addition, statistical significance was evaluated using the Wilcoxon signed-rank test, and p-values were reported to determine whether the performance differences between the proposed method and comparison methods were statistically significant.

### 4.3. The Performance of Our Proposed Approach

In this section, we benchmark the proposed model against state-of-the-art general deep convolutional neural networks and carotid artery ultrasound segmentation-specialized models on the Common Carotid Artery Ultrasound Image Dataset. Chen et al. [41] proposed DeepLab v3+ to fuse multi-scale contextual information with fine spatial details. Isensee et al. introduced the self-configuring nnU-Net, which adapts architectures and training strategies to diverse datasets. Chen et al. [42] developed TransUNet, combining Transformer’s global self-attention with U-Net’s local spatial recovery capability. Ibtehaz et al. [43] proposed MultiResUNet, an improved U-Net with MultiRes blocks and Res paths. Wazir and Kim [44] designed MCADS-Net, featuring a novel MCADS (Multiscale Convolution Attention with Depth-to-Space) decoder and modified U2-Net encoder. Lei et al. [45] presented ConDSeg, addressing soft boundaries via consistency reinforcement and semantic information decoupling modules. Ottakath et al. [19] proposed DoubleUNet-BAM with a Bottleneck Attention Module (BAM) for dual-attention feature capture. Jiang et al. [46] developed MHAHF-UNet, using median enhanced orthogonal convolution to suppress ultrasound noise. Li et al. [47] designed FRDD-Net, integrating feature remapping modules with a dense decoding mechanism. All these methods were re-implemented under consistent experimental settings to ensure a fair comparison.

**Tab. 1** summarizes the quantitative comparison. Our proposed method achieved a Dice Similarity Coefficient (DSC) score of 96.46%, outperforming all competing approaches. Specifically, among general-purpose models, DeepLab v3+, nnU-Net, TransUNet, MultiResUNet, MCADS-Net, and ConDSeg yielded scores of 96.00%, 96.08%, 95.64%, 96.16%, 95.45%, and 96.39%, respectively. Among the specialized models, DoubleUNet-BAM, MHAHF-UNet, and FRDD-Net achieved DSC of 95.89%, 96.26%, and 96.20%, respectively. In addition to region-overlap performance, the proposed method obtained the lowest HD95 value of 2.10 ± 0.13, indicating smaller boundary deviation and more accurate contour localization. **Fig. 6** illustrates the loss and DSC curves from the 5-fold cross-validation. During training, the loss consistently decreased while the DSC improved across all folds, indicating effective learning of discriminative features. On the held-out fold, the loss initially declined and then stabilized, while the DSC score rose before plateauing. These trends demonstrate no signs of overfitting, confirming the model's robustness and stability. Moreover, the statistical analysis yielded p-values lower than 0.05 for all comparisons, confirming that the improvements introduced by the proposed method are statistically significant rather than incidental.

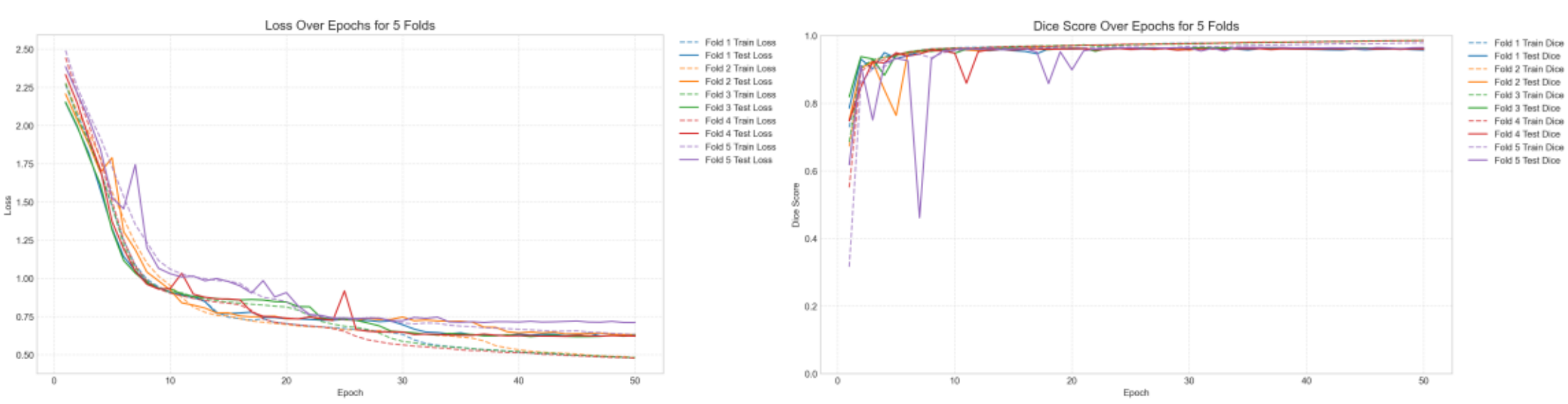

**Fig. 6** Training and held-out-fold loss/DSC curves of FSS-Net across the five folds.

**Fig. 7** presents qualitative segmentation results on the held-out fold. As observed, the model accurately delineates carotid lumen contours and preserves good contour continuity in most cases. Even in regions with blurred vessel-wall boundaries caused by limited resolution or tissue intensity overlap, the model can still produce relatively complete lumen segmentation results. These qualitative results suggest that the proposed method can provide a useful segmentation basis for subsequent carotid ultrasound morphology analysis. These results demonstrate that the proposed model excels not only in typical scenarios but also in complex lesion segmentation, providing a reliable foundation for downstream quantitative tasks such as plaque volume measurement and stability assessment.

**Tab. 1** Performance comparison of the Common Carotid Artery Ultrasound Images Dataset.

| Method | DSC | IoU | Precision | Recall | Accuracy | Spec | MCC | Parameters | HD95 | P-value |
|---|---|---|---|---|---|---|---|---|---|---|
| U-Net [36] | 96.08±0.19 | 92.65±0.34 | 96.28±0.43 | 96.12±0.32 | 99.82±0.01 | 99.91±0.01 | 96.10±0.19 | **1.94 M** | **2.60±0.31** | $2.65\times10^{-4}$ |
| DeepLab v3+ [41] | 96.00±0.15 | 92.32±0.28 | 95.76±0.66 | 96.26±0.52 | 99.81±0.01 | 99.90±0.02 | 95.91±0.16 | 41.99 M | 2.57±0.43 | $1.22\times10^{-4}$ |
| nnU-Net [48] | 96.30±0.13 | 92.87±0.24 | 96.30±0.29 | 96.31±0.26 | 99.83±0.01 | 99.91±0.01 | 96.21±0.14 | 9.24 M | 2.64±0.60 | $1.62\times10^{-2}$ |
| TransUNet [42] | 95.64±0.32 | 91.67±0.57 | 96.03±0.40 | 95.32±0.90 | 99.80±0.02 | 99.91±0.01 | 95.55±0.32 | 8.76 M | 3.51±0.52 | $5.76\times10^{-5}$ |
| MultiResUNet [43] | 96.16±0.17 | 92.62±0.20 | 96.10±0.17 | 96.24±0.23 | 99.82±0.01 | 99.91±0.01 | 96.07±0.17 | 8.02 M | 2.28±0.08 | $4.04\times10^{-4}$ |
| MCADS-Net [44] | 95.45±0.18 | 91.33±0.32 | 95.40±0.35 | 95.58±0.34 | 99.79±0.01 | 99.89±0.01 | 95.36±0.19 | 95.20 M | 2.47±0.07 | $4.04\times10^{-5}$ |
| ConDSeg [45] | 96.39±0.16 | 93.03±0.30 | 96.39±0.41 | 96.40±0.35 | 99.83±0.01 | 99.91±0.01 | 96.30±0.16 | 45.50 M | 2.26±0.35 | $4.11\times10^{-2}$ |
| DoubleUNet-BAM [19] | 96.06±0.35 | 92.44±0.63 | 96.00±0.48 | 96.14±0.38 | 99.82±0.02 | 99.90±0.01 | 95.97±0.36 | 36.72 M | 2.54±0.42 | $2.87\times10^{-4}$ |
| MHAHF-UNet [46] | 96.26±0.16 | 92.81±0.29 | 95.87±0.27 | **96.68±0.23** | 99.18±0.01 | 99.92±0.01 | 95.36±0.15 | 9.26 M | 2.53±0.67 | $5.73\times10^{-4}$ |
| FRDD-Net [47] | 96.20±0.13 | 92.69±0.23 | 96.48±0.55 | 95.96±0.37 | 99.82±0.01 | 99.92±0.01 | 96.12±0.13 | 3.11 M | 2.72±0.23 | $6.09\times10^{-4}$ |
| **Ours** | **96.46±0.13** | **93.18±0.24** | **96.46±0.27** | <u>96.50±0.42</u> | **99.84±0.01** | **99.92±0.02** | **96.55±0.13** | 5.01 M | **2.10±0.13** | - |

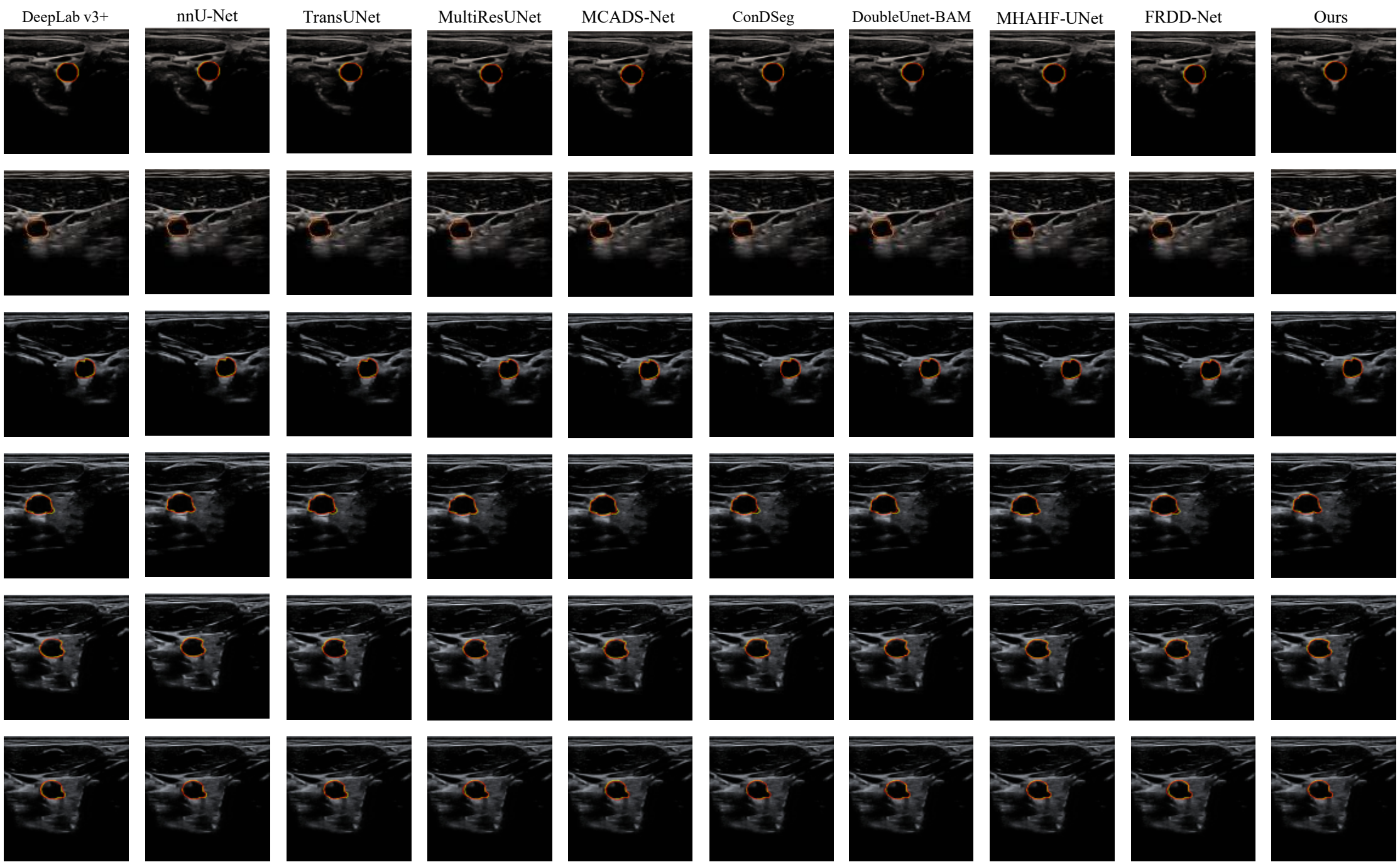

**Fig. 7** Segmentation results of various models. The yellow line denotes the ground truth region, and the red line denotes the predicted region.

### 4.4. Ablation Study

To systematically validate the individual contributions of each component within the proposed framework, ablation studies were conducted on the Common Carotid Artery Ultrasound Image Dataset. As detailed in **Tab. 2**, the standard U-Net was employed as the baseline, with the DSC serving as the primary evaluation metric.

Initially, we integrated each module independently into the baseline U-Net to evaluate its specific impact. The baseline model yielded a DSC of 96.08% and an IoU of 92.65%. Incorporating the Channel-Spatial-Wavelet Attention (CSWA), Laplacian-Guided Adaptive Edge Fusion (LAEF), and Wavelet-Enhanced Bottleneck (WEB) modules individually improved the DSC to 96.30%, 96.37%, and 96.20%, respectively.

Subsequently, to verify the necessity of each component within the complete framework, we performed exclusion experiments by systematically removing specific modules from the final model (Ours). Specifically, excluding the CSWA module (U-Net + LAEF + WEB) caused the DSC to drop from 96.46% to 96.28%. When substituting the LAEF module with standard skip connections (U-Net + CSWA + WEB), the DSC score decreased to 96.38%. Finally, replacing the WEB with a conventional bottleneck layer (U-Net + CSWA + LAEF) resulted in a decline to 96.39%.

To clarify whether the performance improvement mainly originates from the proposed architectural design or from a simple cumulative effect of engineering components, we conducted an architecture–loss disentanglement ablation study. As shown in **Tab. 3**, adding Boundary Loss to the standard U-Net resulted in almost no improvement in DSC, IoU, or MCC. This indicates that, when the network lacks explicit boundary-aware feature modeling, Boundary Loss alone is insufficient to produce a clear performance gain. In contrast, FSS-Net trained only with BCE + DSC achieved a DSC of 96.37±0.14 and an MCC of 96.25±0.15. After introducing Boundary Loss, the full model further improved to 96.46 ±0.13 in DSC, 93.18±0.24 in IoU, and 96.55±0.13 in MCC. The results in **Tab. 2** and **Tab. 3** indicate that the performance gain is not merely caused by stacking an additional loss term.

**Tab. 2** Ablation study of different components in FSS-Net on the Common Carotid Artery Ultrasound Dataset.

| Method | DSC | IoU | Precision | Recall | Accuracy | Spec | MCC |
|---|---|---|---|---|---|---|---|
| U-Net | 96.08±0.19 | 92.65±0.34 | 96.28±0.43 | 96.12±0.32 | 99.82±0.01 | 99.91±0.01 | 96.10±0.19 |
| U-Net + CSWA | 96.30±0.16 | 92.87±0.29 | 96.20±0.34 | 96.44±0.38 | 99.83±0.01 | 99.91±0.01 | 96.22±0.16 |
| U-Net + LAEF | 96.37±0.15 | 93.09±0.27 | 96.27±0.53 | 96.59±0.49 | 99.83±0.01 | 99.91±0.01 | 96.34±0.15 |
| U-Net + WEB | 96.20±0.21 | 92.67±0.37 | 96.32±0.47 | 96.10±0.44 | 99.82±0.01 | 99.91±0.01 | 96.10±0.21 |
| U-Net + LAEF + WEB | 96.28±0.09 | 92.86±0.20 | 95.92±0.28 | 96.70±0.35 | 99.82±0.01 | 99.90±0.01 | 96.21±0.13 |
| U-Net + CSWA + WEB | 96.38±0.10 | 93.02±0.18 | 96.41±0.46 | 96.37±0.45 | 99.83±0.01 | 99.91±0.01 | 96.30±0.10 |
| U-Net + CSWA + LAEF | 96.39±0.14 | 93.04±0.25 | 96.34±0.25 | 96.47±0.29 | 99.83±0.01 | 99.91±0.01 | 96.31±0.14 |
| **Ours** | **96.46±0.13** | **93.18±0.24** | **96.46±0.27** | **96.50±0.42** | **99.84±0.01** | **99.92±0.02** | **96.55±0.13** |

**Tab. 3** Ablation study of architectural design and loss components on the Common Carotid Artery Ultrasound Dataset.

| Method | DSC | IoU | Precision | Recall | Accuracy | Spec | MCC |
|---|---|---|---|---|---|---|---|

| U-Net + BCE + DSC | 96.08±0.16 | 92.65±0.29 | 96.37±0.38 | 95.84±0.52 | 99.82±0.01 | 99.92±0.01 | 96.10±0.16 |
|---|---|---|---|---|---|---|---|
| U-Net + BCE + DSC + Boundary Loss | 96.08±0.19 | 92.65±0.34 | 96.28±0.43 | 96.12±0.32 | 99.82±0.01 | 99.91±0.01 | 96.10±0.19 |
| FSS-Net + BCE + DSC | 96.37±0.14 | 92.98±0.27 | 96.34±0.32 | **96.54±0.56** | 99.83±0.01 | 99.91±0.01 | 96.25±0.15 |
| **Ours** | **96.46±0.13** | **93.18±0.24** | **96.46±0.27** | 96.50±0.42 | **99.84±0.01** | **99.92±0.02** | **96.55±0.13** |

To further substantiate the efficacy of the CSWA module, we benchmarked it against two well-established attention mechanisms: the Bottleneck Attention Module (BAM) and the Convolutional Block Attention Module (CBAM). Like CSWA, these are hybrid mechanisms that synergize spatial and channel attention. However, our proposed method demonstrated superior performance across critical metrics, including DSC, IoU, Precision, and MCC. As detailed in **Tab. 4**, BAM and CBAM achieved DSC of 96.36% and 96.14%, respectively—both of which are inferior to the results obtained by our proposed CSWA module.

**Tab. 4** Comparative experiment of BAM, CBAM, and CSWA on Common Carotid Artery Ultrasound Images Dataset.

| Method | DSC | IoU | Precision | Recall | Accuracy | Spec | MCC |
|---|---|---|---|---|---|---|---|
| BAM [25] | 96.36±0.12 | 92.98±0.22 | 96.13±0.31 | **96.61±0.30** | 99.83±0.01 | 99.91±0.01 | 96.27±0.12 |
| CBAM [26] | 96.14±0.16 | 92.58±0.30 | 95.84±0.69 | 96.47±0.47 | 99.82±0.01 | 99.90±0.01 | 96.06±0.17 |
| **CSWA (ours)** | **96.46±0.13** | **93.18±0.24** | **96.46±0.27** | 96.50±0.42 | **99.84±0.01** | **99.92±0.02** | **96.55±0.13** |

To evaluate the efficacy and efficiency of the WEB module in feature extraction, we benchmarked it against two classic bottleneck designs: the Spatial Attention Module (SAM) [26] and Atrous Spatial Pyramid Pooling (ASPP) [41]. **Tab. 5** details the segmentation performance metrics for each method. The results indicate that the WEB achieved superior performance across key metrics, including DSC (96.46%), IoU (93.18%), and MCC (96.55%). It outperformed the computationally intensive ASPP scheme (96.42% DSC) and significantly surpassed SAM (96.37% DSC). Regarding model complexity, although ASPP expands the receptive field through multi-path dilated convolutions, it incurs a substantial computational cost of 554.04 M MACs and 2.23 M parameters. In contrast, leveraging the efficiency of the wavelet transform, the WEB consumes only 134.30 M MACs and 0.53 M parameters. By achieving higher segmentation accuracy at approximately one-quarter of the computational cost of ASPP, our module demonstrates a compelling balance between capturing fine-grained features and maintaining computational efficiency.

**Tab. 5** Comparative experiment of SAM, ASPP, and WEB on Common Carotid Artery Ultrasound Images Dataset.

| Method | DSC | IoU | Precision | Recall | Accuracy | Spec | MCC |
|---|---|---|---|---|---|---|---|
| SAM [26] | 96.37±0.17 | 93.01±0.32 | 96.26±0.40 | 96.50±0.48 | 99.83±0.01 | 99.91±0.01 | 96.26±0.18 |
| ASPP [41] | 96.42±0.14 | 93.17±0.27 | 96.30±0.41 | **96.64±0.35** | 99.83±0.01 | 99.90±0.01 | 96.38±0.15 |
| **WEB (ours)** | **96.46±0.13** | **93.18±0.24** | **96.46±0.27** | 96.50±0.42 | **99.84±0.01** | **99.92±0.02** | **96.55±0.13** |

Additionally, to validate the effectiveness of the LAEF module, comparative experiments were performed against the classic Attention Gate (AG) [49] mechanism and ResBlock [50]. The results are presented in **Tab. 6**.

**Tab. 6** Comparative experiment of AG, ResBlock, and LAEF on Common Carotid Artery Ultrasound Images Dataset.

| Method | DSC | IoU | Precision | Recall | Accuracy | Spec | MCC |
|---|---|---|---|---|---|---|---|
| AG [49] | 96.25±0.18 | 92.77±0.32 | 96.43±0.21 | 96.08±0.41 | 99.82±0.01 | 99.91±0.01 | 96.26±0.18 |
| ResBlock [50] | 96.37±0.13 | 93.01±0.24 | 96.50±0.51 | 96.27±0.46 | 99.83±0.01 | 99.92±0.01 | 96.29±0.13 |
| **LAEF (ours)** | **96.46±0.13** | **93.18±0.24** | **96.46±0.27** | **96.50±0.42** | **99.84±0.01** | **99.92±0.02** | **96.55±0.13** |

The results demonstrate that the LAEF module exhibits superior performance compared to the competing modules. Specifically, LAEF achieved a DSC of 96.46% and an IoU of 93.18%, surpassing both AG (96.25%, 92.77%) and ResBlock (96.37%, 93.01%). Moreover, LAEF maintained a consistent advantage in terms of Recall, Accuracy, and MCC. These findings indicate that the LAEF module is more effective in capturing critical feature information, thereby significantly enhancing the segmentation accuracy of carotid artery ultrasound images.

To further analyze the contribution of different frequency sub-bands, we conducted a sub-band ablation study on CSWA. Since the LL sub-band provides the main low-frequency structural representation, it was retained in all variants. We then removed one high-frequency sub-band at a time, including LH, HL, and HH, to evaluate the contribution of directional detail information. As shown in **Tab. 7**, removing any high-frequency sub-band led to a decrease in the overall segmentation performance compared with the full setting. The complete configuration using LL+LH+HL+HH achieved the best DSC of 96.46 ± 0.13, IoU of 93.18 ± 0.24, and MCC of 96.55 ± 0.13. These results indicate that the high-frequency sub-bands are not merely noise components; instead, they provide complementary directional details that contribute to boundary preservation and region discrimination.

**Tab. 7** Sub-band ablation study on the Common Carotid Artery Ultrasound Images Dataset.

| Method | DSC | IoU | Precision | Recall | Accuracy | Spec | MCC |
|---|---|---|---|---|---|---|---|
| LL+HL+HH | 96.38±0.11 | 93.13±0.19 | 96.42±0.44 | 96.57±0.30 | 99.84±0.01 | 99.92±0.01 | 96.41±0.11 |
| LL+LH+HH | 96.29±0.29 | 92.98±0.44 | **96.59±0.38** | 96.21±0.32 | 99.83±0.01 | 99.92±0.01 | 96.25±0.27 |
| LL+LH+HL | 96.35±0.16 | 93.03±0.28 | 96.18±0.31 | **96.67±0.31** | 99.83±0.01 | 99.91±0.01 | 96.30±0.16 |
| LL+LH+HL+HH | **96.46±0.13** | **93.18±0.24** | 96.46±0.27 | 96.50±0.42 | **99.84±0.01** | **99.92±0.02** | **96.55±0.13** |

### 4.5. Robustness of Our Approach to Noise

Speckle noise is an intrinsic artifact in ultrasound imaging, arising from the interference of backscattered coherent waves. This granular noise typically manifests as multiplicative contamination, which degrades high-frequency edge details and complicates the segmentation task. To rigorously assess the model's stability in realistic clinical scenarios, this study applied a Multiplicative Speckle Noise Model [51] to the test set. The specific mathematical formulation is expressed as follows:

$$N(x,y) = g(x,y) \cdot [1 + u(x,y)\,] \tag{16}$$

where $g(x,y)$ represents the original image, $u(x,y)$ represents Gaussian noise with a mean of 0 and variance $s$, which is related to the grayscale values of the original image: the higher the grayscale value, the higher the variance of the noise. $N(x,y)$ denotes the noisy image.

To simulate varying degrees of image degradation, speckle noise was superimposed onto the Common Carotid Artery Ultrasound Images dataset. The severity of the added noise is quantified using the Peak Signal-to-Noise Ratio (PSNR), where a lower PSNR value corresponds to a higher noise intensity. The PSNR is defined mathematically as:

$$\mathrm{PSNR} = 20 \cdot \log_{10}\left(\frac{Max}{\sqrt{MSE}}\right) \tag{17}$$

where $Max$ denotes the maximum possible pixel intensity of the image, and MSE represents the Mean Squared Error between the noise-corrupted image and the original clean image. To rigorously assess robustness, models trained exclusively on noise-free data were directly evaluated on this noisy test set to measure performance stability under varying noise conditions.

To objectively evaluate noise robustness, speckle noise (ranging from 28.82 dB to 19.28 dB) was introduced exclusively to the test set of the Common Carotid Artery Ultrasound Image Dataset; crucially, all models were trained solely on data without additional noise. As shown in **Tab. 8**, as noise intensity increases (i.e., as the Peak Signal-to-Noise Ratio (PSNR) decreases), the performance of most comparative models degraded to varying degrees. In contrast, our method exhibited exceptional noise robustness.

**Tab. 8** Comparison of noise robustness across models on Common Carotid Artery Ultrasound Images Dataset.

| Method \ Noise intensity | 28.82 dB | 25.46 dB | 23.13 dB | 21.46 dB | 20.22 dB | 19.28 dB |
|---|---|---|---|---|---|---|
| DeepLab v3+ [41] | 91.48±4.88 | 91.46±4.94 | 91.43±4.95 | 91.41±5.02 | 91.33±5.08 | 91.16±5.28 |
| nnU-Net [48] | **94.78±1.07** | **94.78±1.07** | **94.79±1.07** | **94.81±1.06** | **94.81±1.06** | **94.83±1.04** |
| TransUNet [42] | 87.51±9.16 | 87.48±9.19 | 87.44±9.19 | 87.43±9.17 | 87.32±9.31 | 87.21±9.32 |
| MCADS-Net [44] | 51.20±24.70 | 51.16±22.73 | 51.65±23.88 | 51.56±24.27 | 52.04±24.33 | 52.12±24.39 |
| ConDSeg [45] | 85.51±9.78 | 84.91±10.25 | 84.51±10.26 | 84.29±10.08 | 84.01±10.40 | 83.82±10.25 |
| DoubleUNet-BAM [19] | 62.53±14.00 | 63.19±13.82 | 63.82±13.67 | 64.23±13.68 | 64.63±13.75 | 64.65±13.64 |
| MultiResUNet [43] | 66.68±18.33 | 66.61±18.31 | 66.49±18.45 | 66.17±18.55 | 65.71±18.65 | 65.26±18.88 |
| MHAHF-UNet [46] | 92.22±1.72 | 92.22±1.72 | 92.20±1.73 | 92.17±1.75 | 92.13±1.76 | 92.09±1.80 |
| FRDD-Net [47] | 91.93±3.80 | 91.94±3.74 | 91.95±3.61 | 91.88±3.67 | 91.89±3.61 | 91.79±3.90 |
| Ours | 92.39±3.04 | 92.36±3.03 | 92.34±3.08 | 92.31±3.01 | 92.27±3.14 | 92.19±3.04 |

Specifically, under the highest noise intensity (19.28 dB), the performance of DeepLab v3+ dropped from 91.48% to 91.16%, TransUNet declined from 87.51% to 87.21%, and ConDSeg decreased from 85.51% to 83.82%. In comparison, although some models such as MCADS-Net, DoubleUNet-BAM, and MultiResUNet appear numerically stable, they exhibit extremely high standard deviations (ranging from approximately ±13% to ±24.7%). This suggests that their predictions are highly unstable and lack reliability in noisy environments. In sharp contrast, our method maintained highly stable performance across all noise levels. Under minimal noise interference (28.82 dB), our model achieved a performance score of 92.39%. Even under the most severe noise conditions (19.28 dB), the performance experienced only a marginal fluctuation to 92.19%, representing a negligible decline of just 0.2%. Simultaneously, our model consistently maintained a low standard deviation (approximately ±3.04%), further verifying the consistency and reliability of its predictions.

While nnU-Net demonstrated strong noise robustness and slightly higher absolute metrics, this performance comes at the cost of a substantial parameter count of 9.24 M. In sharp contrast, our method achieves robust noise resilience with only 5.01 M parameters—approximately half that of nnU-Net—while significantly outperforming other mainstream networks such as DeepLab v3+, TransUNet, and FRDD-Net. This indicates that the proposed network architecture effectively extracts key features in both the frequency and spatial domains, thereby suppressing noise interference while accurately preserving anatomical structural details crucial for segmentation tasks. These experimental results fully validate the robustness of our method in complex noisy environments, highlighting its significant value for practical clinical applications.

**Tab. 9** Robustness comparison of noise-trained models on the Common Carotid Artery Ultrasound Images Dataset.

| Method \ Noise intensity | 28.82 dB | 25.46 dB | 23.13 dB | 21.46 dB | 20.22 dB | 19.28 dB |
|---|---|---|---|---|---|---|
| DeepLab v3+ [41] | 95.99±0.16 | 95.87±0.20 | 95.82±0.17 | 95.74±0.16 | 95.67±0.15 | 95.57±0.31 |
| nnU-Net [48] | 96.16±0.16 | 96.14±0.15 | 96.07±0.11 | 96.03±0.16 | 95.96±0.19 | 95.92±0.18 |
| TransUNet [42] | 95.65±0.05 | 95.37±0.23 | 95.50±0.29 | 95.30±0.30 | 95.44±0.39 | 95.38±0.21 |
| MCADS-Net [44] | 95.39±0.32 | 95.33±0.18 | 95.25±0.14 | 95.20±0.11 | 95.12±0.18 | 95.09±0.12 |
| ConDSeg [45] | 96.11±0.15 | 96.06±0.25 | 96.00±0.17 | 95.91±0.09 | 95.80±0.16 | 95.76±0.17 |
| DoubleUNet-BAM [19] | 96.23±0.17 | 96.03±0.21 | 95.94±0.28 | 96.00±0.24 | 96.10±0.24 | 95.90±0.17 |
| MultiResUNet [43] | 96.09±0.20 | 95.95±0.25 | 95.88±0.18 | 95.81±0.25 | 95.82±0.22 | 95.77±0.21 |
| MHAHF-UNet [46] | 96.21±0.21 | 96.06±0.22 | 95.97±0.18 | 95.93±0.17 | 95.91±0.14 | 95.85±0.19 |
| FRDD-Net [47] | 96.15±0.17 | 96.00±0.20 | 95.96±0.12 | 95.86±0.20 | 95.81±0.18 | 95.85±0.13 |
| Ours | **96.31±0.12** | **96.19±0.09** | **96.10±0.11** | **96.09±0.19** | **95.99±0.19** | **95.94±0.13** |

To further evaluate the robustness of different models under noisy training conditions, we conducted additional experiments in which speckle noise was added to the dataset at six fixed PSNR levels, including 28.82 dB, 25.46 dB, 23.13 dB, 21.46 dB, 20.22 dB, and 19.28 dB. For each noise level, all models were independently trained and evaluated using five-fold cross-validation. As shown in **Tab. 9**, the proposed method achieved the highest average DSC under all noise intensities. Specifically, our method obtained DSC values of 96.31%, 96.19%, 96.10%, 96.09%, 95.99%, and 95.94% from mild to severe noise levels, respectively. Although the segmentation performance gradually decreased as the noise intensity increased, the proposed method maintained stable results with relatively small standard deviations. Compared with other competitive methods, such as nnU-Net, MHAHF-UNet, and FRDD-Net, our model consistently showed better average performance across different noise levels. These results indicate that the proposed frequency-spatial design can effectively adapt to noisy ultrasound images when trained under corresponding noise conditions, further supporting its robustness in low-SNR scenarios.

### 4.6. Assessment of Model Generalization Capability

To further verify the generalization ability of the method proposed in this paper across different anatomical structures and imaging scenarios, in addition to the primary carotid artery dataset, we also conducted extended experiments on the Breast Ultrasound Images Dataset. The proposed model was compared with MultiResUNet, TransUNet, MCADS-Net, ConDSeg, MHAHF-UNet, and EU2-Net. According to the quantitative results presented in **Tab. 10**, our model still performs excellently even on breast ultrasound images with morphological characteristics distinct from those of carotid arteries. Specifically, our method achieved the highest DSC of 75.86% and competitive IoU performance on the BUSI dataset, significantly outperforming comparative methods such as MHAHF-UNet. This result strongly demonstrates that our model not only exhibits outstanding performance on the carotid artery segmentation task but also possesses favorable robustness and adaptability when handling other types of medical ultrasound image segmentation tasks.

**Tab. 10** Performance comparison of the Breast Ultrasound Images Dataset.

| Method | DSC | Recall | Precision | IoU | Specificity |
|---|---|---|---|---|---|
| MultiResUNet [43] | 73.45±2.93 | 70.60±2.16 | 79.32±4.55 | 59.34±2.99 | **98.12±0.61** |
| TransUNet [42] | 65.99±3.73 | 68.14±6.03 | 66.88±3.82 | 50.38±3.81 | 96.47±1.13 |
| FRDD-Net [47] | 71.06±2.68 | 70.27±4.02 | 73.72±3.21 | 56.50±2.66 | 97.17±0.63 |
| MCADS-Net [44] | 71.80±5.47 | 73.03±2.84 | 70.53±6.45 | 54.70±5.87 | 96.96±1.01 |
| ConDSeg [45] | 72.60±3.18 | 75.75±2.26 | 70.76±4.08 | 57.67±3.70 | 96.83±0.63 |
| MHAHF-UNet [46] | 73.86±4.14 | 72.53±3.53 | 76.10±4.64 | 59.28±4.83 | 97.53±0.63 |
| EU2-Net [52] | 75.55±4.14 | **79.71±3.25** | **78.27±5.31** | **65.66±5.24** | 97.66±0.66 |
| **Ours** | **75.86±3.77** | 74.67±4.46 | 78.13±3.34 | 61.97±4.31 | 97.79±0.26 |

## 5. Discussion

### 5.1. The Performance of FSS-Net

The performance superiority of FSS-Net in carotid artery ultrasound segmentation stems from a fundamental architectural advantage. Rather than being hampered by the physical limitations of standard convolutions, FSS-Net addresses the core bottleneck of existing methods through an innovative frequency-spatial synergistic design, fundamentally shifting the paradigm of ultrasound image processing.

The primary cause of underperformance in competing methods (e.g., nnU-Net, DeepLab v3+, TransUNet) is spatial entanglement. Most state-of-the-art models are restricted to spatial-domain feature extraction. In carotid ultrasound imaging—characterized by blurred vessel walls, plaque heterogeneity, and acoustic artifacts—multiplicative speckle noise and anatomical boundaries exhibit overlapping intensity distributions. Because spatial convolutions rely strictly on these pixel intensities, they cannot effectively distinguish between high-frequency noise clusters and structural edges. This entanglement forces an inherent trade-off between aggressive smoothing and detail preservation, ultimately resulting in the performance degradation observed in **Tab. 1**.

In stark contrast, the fundamental driver of FSS-Net's success is spectral separation, realized through a cohesive frequency-spatial synergistic pipeline. By integrating wavelet transforms, the network physically isolates noise components into specific high-frequency sub-bands. Specifically, the Channel-Spatial-Wavelet Attention (CSWA) module first disentangles noise in the encoder to establish a robust semantic foundation; subsequently, the Wavelet-Enhanced Bottleneck (WEB) module leverages these refined features at the bottleneck to model global long-range dependencies without artifact interference; finally, the Laplacian-Guided Adaptive Edge Fusion (LAEF) module actively compensates for potential detail loss during spectral filtering by explicitly restoring boundary topology in the decoder. This sequential design robustly resolves the spatial entanglement dilemma. As evidenced by the ablation experiments (**Tab. 2**), removing any component disrupts this delicate equilibrium, confirming that the holistic integration of these core modules is critical for achieving state-of-the-art segmentation performance.

The results in **Tab. 2** and **Tab. 3** further demonstrate that the improvement of FSS-Net cannot be simply attributed to the cumulative effect of stacked components. **Tab. 3** shows that Boundary Loss alone produced limited improvement on the standard U-Net, but became more effective when combined with the proposed CSWA, WEB, and LAEF modules. Consistent with the component-wise ablation results in **Tab. 2**, this observation suggests that the final performance gain mainly arises from the complementary interaction among frequency-aware encoding, bottleneck-level contextual modeling, decoder-stage edge refinement, and boundary-aware supervision.

Quantitative results in **Tab. 1** verify that FSS-Net consistently outperforms both general medical image segmentation models and specialized carotid artery segmentation models across key metrics, including DSC, IoU, and MCC. For intuitive and interpretable evaluation, Grad-CAM-generated heatmaps (**Fig. 8**) demonstrate that FSS-Net accurately localizes and fully covers the carotid artery region with high attentional alignment, while competing models either fail to focus effectively on the target region or exhibit overly narrow attention on boundary details. Notably, **Tab. 8** confirms FSS-Net's excellent robustness against speckle noise, a direct benefit of its spectral separation mechanism, which further distinguishes it from spatial-domain-dominant competitors.

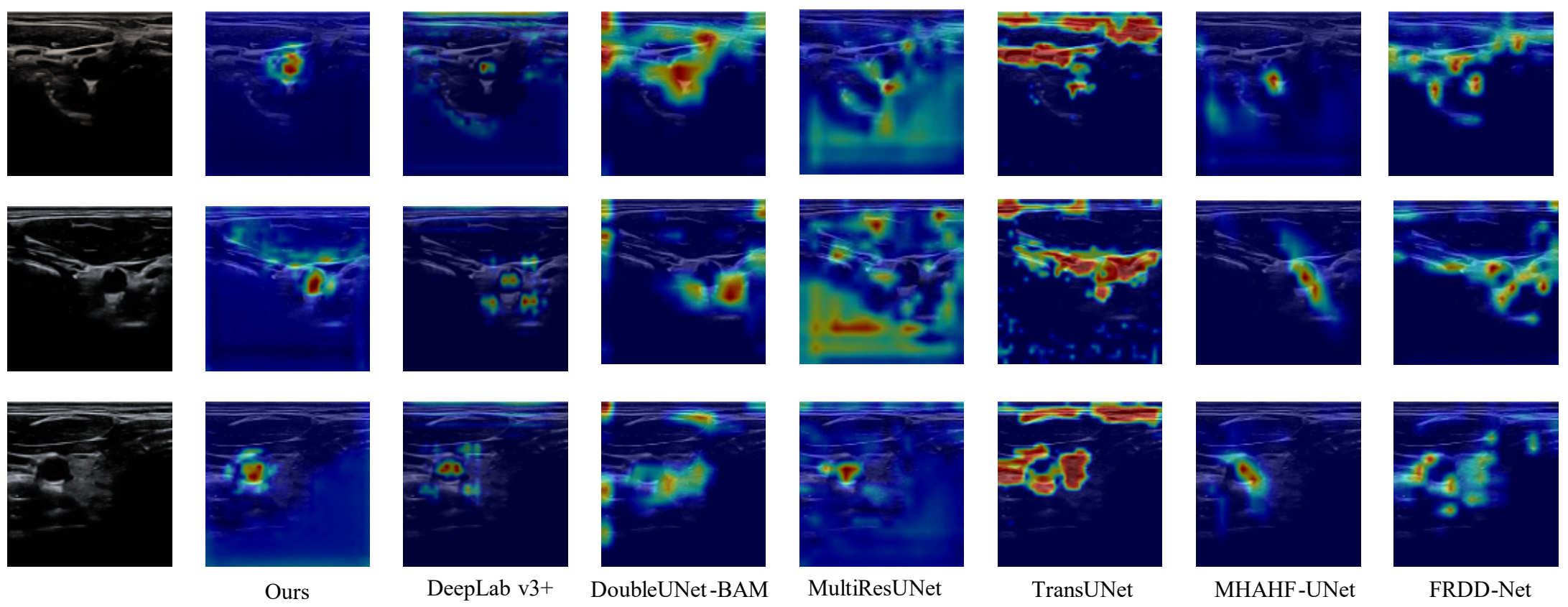

**Fig. 8** Heatmaps of Common Carotid Artery Ultrasound Images

In terms of model complexity, FSS-Net's parameter count (5.01 M) is moderately increased compared to the standard U-Net and FRDD-Net, well within the acceptable range for clinical deployment. In contrast, nnU-Net—though slightly more robust to speckle noise—has nearly double the parameter count of FSS-Net yet still underperforms on core segmentation tasks (**Tab. 1**). Collectively, the results from **Tab. 1**, **Tab. 2**, and **Tab. 8**, combined with qualitative evidence from **Fig. 8**, confirm that FSS-Net's frequency-domain prior-based design provides a fundamental solution to ultrasound artifacts, offering clear advantages over pure spatial-domain models.

### 5.2. Efficacy of Attention Modules

The comparative analysis against BAM [25] and CBAM [26] highlights the robustness gap between purely spatial attention and frequency-aware mechanisms. **Fig. 9** reveals that CBAM suffers from spatial myopia: its fixed kernel size limits the effective receptive field, resulting in fragmented activations that fail to cover the full vessel topology. Meanwhile, BAM's strategy of using dilated convolutions backfires in noisy environments. The discrete sampling of dilated kernels creates spatial discontinuities (gridding artifacts), causing the network to mistake high-frequency speckle noise for salient features, thereby introducing substantial background noise into the attention map.

To alleviate the limitations of purely spatial feature modulation, the proposed CSWA module introduces joint channel, spatial, and wavelet-domain attention. Instead of relying only on spatial responses, CSWA decomposes feature maps into wavelet sub-bands before attention generation. The low-frequency approximation coefficients mainly encode coarse anatomical structures, while the high-frequency detail coefficients capture direction-sensitive responses that include both boundary-related details and noise-sensitive local fluctuations. In this way, DWT serves as an interpretable feature decomposition operation rather than a hard noise-separation mechanism. By applying attention weights across this decoupled spectrum prior to spatial reconstruction, CSWA actively filters out artifacts at the source. The sub-band ablation in **Tab. 7** further supports this interpretation: retaining all high-frequency sub-bands together with LL-guided modulation achieved the best overall performance, suggesting that LH, HL, and HH contain useful boundary-related details in addition to noise-sensitive fluctuations. This frequency-spatial synergy ensures that the network selectively amplifies true vascular boundaries and plaque morphologies, achieving high-precision localization without suffering from the gridding artifacts or structural distortions typical of purely spatial dilation techniques.

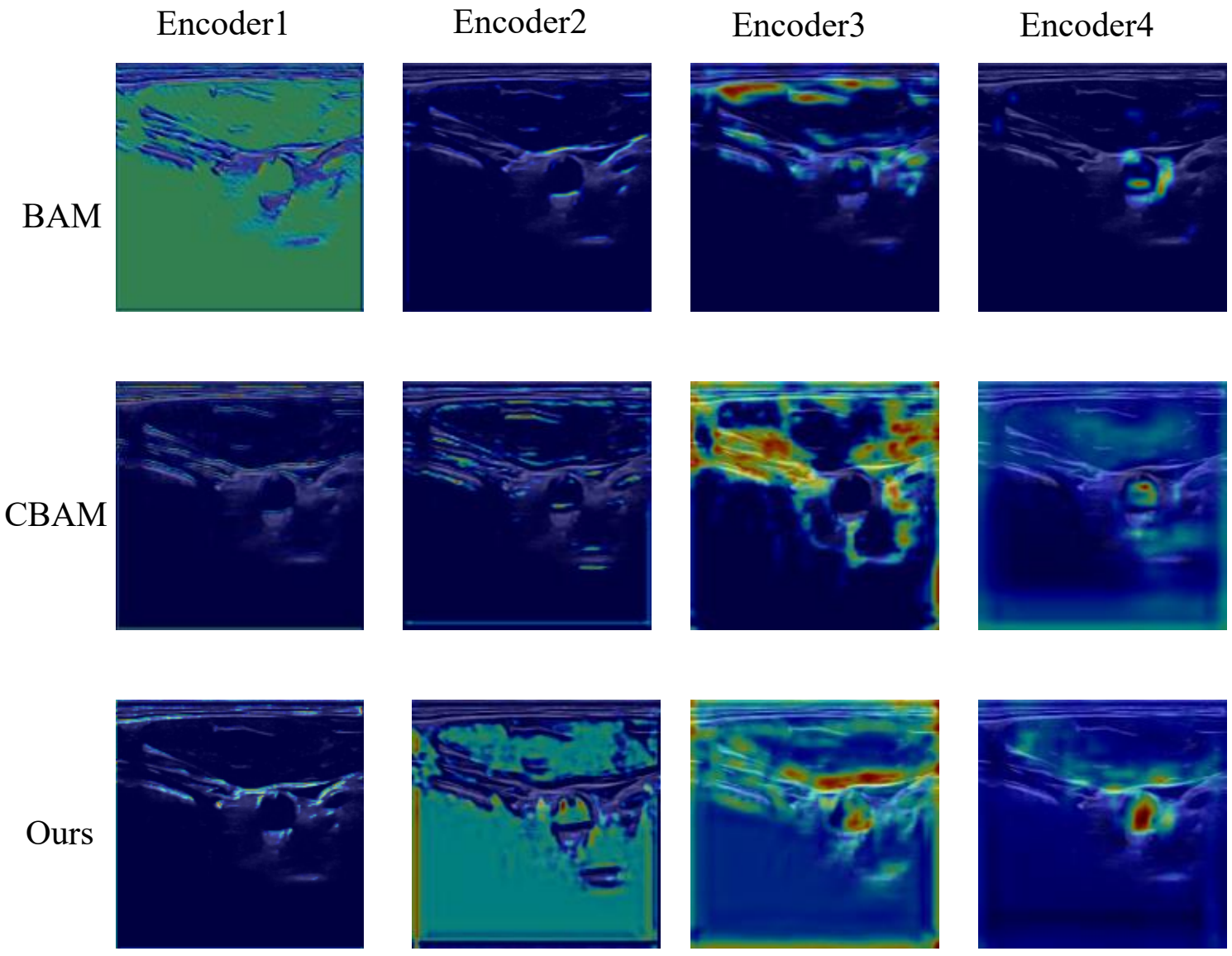

**Fig. 9** Visualization analysis of BAM, CBAM and CSWA

### 5.3. Efficacy of Edge Enhancement and Attention Modules

The ablation studies detailed in Section 4.4 empirically validate the critical role of the Laplacian-Guided Adaptive Edge Fusion (LAEF) module in resolving the ambiguous boundary challenge inherent in carotid ultrasound imaging. Quantitatively, mitigating the semantic gap between the encoder and the decoder using the LAEF module yielded substantial improvements in Recall and MCC, outperforming implicit mechanisms such as Attention Gates (AG) [49] and residual blocks (ResBlock) [50]. This performance gap suggests that traditional skip connections or generic region attention mechanisms struggle to effectively reconstruct fine-grained topological features that are attenuated during continuous downsampling. In contrast, LAEF leverages local context awareness to establish an efficient feature compensation channel, thereby maximally recovering the target's geometric boundaries and topological integrity.

The mechanistic superiority of the LAEF module derives from its unique explicit edge modeling strategy. Rather than employing simple feature superposition, this module extracts high-frequency structural priors via the Laplacian operator and constructs a dynamic complementary mechanism using an adaptive selector. This architecture endows the network with content-aware feature modulation capabilities: it prioritizes reliance on deep semantic representations when boundaries are distinct, while automatically transitioning to a guidance mode dominated by explicit edge cues when boundaries are obscured by artifacts. Furthermore, LAEF effectively compensates for the CSWA module's intrinsic deficiency in capturing high-frequency details. The visualization results in **Fig. 7** corroborate this effectiveness; FSS-Net successfully delineates segmentation contours that are topologically continuous and geometrically accurate, even when processing extremely irregular plaques.

### 5.4. Efficacy of Bottleneck Modules

The bottleneck layer is pivotal for capturing long-range dependencies, yet standard spatial approaches often struggle to balance receptive field expansion with computational efficiency. As evidenced by the comparative experiments in Table IV, while ASPP [41] effectively enlarges the field of view through multipath dilated convolutions, it suffers from intrinsic limitations. The discrete sampling of atrous kernels introduces gridding artifacts, which disrupt the local spatial continuity required for defining precise plaque boundaries. Furthermore, this spatial dilation strategy incurs a substantial computational burden, demanding 554.04 M MACs and 2.23 M parameters to maintain performance. While SAM [26] attempts to address spatial dependencies, its performance remains limited by its spatial-only attention mechanism, failing to fully capture the multiscale heterogeneity of carotid plaques.

In contrast, the WEB module resolves this dilemma by shifting the global modeling paradigm from spatial dilation to spectral compression. By leveraging the Discrete Wavelet Transform (DWT), WEB maps high-dimensional features into compact frequency sub-bands, where the low-frequency component naturally aggregates global structural information. This frequency-domain strategy achieves a superior trade-off: WEB not only attains the highest segmentation accuracy (DSC = 96.46%) but does so with exceptional efficiency. It consumes only 134.30 M MACs and 0.53 M parameters—approximately 25% of the computational load of ASPP. This confirms that exploiting spectral sparsity offers a more robust and lightweight solution for clinical deployment than brute-force spatial expansion.

### 5.5. Noise Robustness via Frequency-Domain Feature Extraction

This study demonstrates that FSS-Net exhibits exceptional stability even in environments characterized by extremely low Signal-to-Noise Ratios (SNR), a capability fully corroborated by the noise robustness experiments presented in **Tab. 8**. While the self-configuring nnU-Net maintains strong resistance to noise interference, its performance relies heavily on a complex architecture with a substantial parameter burden (9.24 M). In contrast, FSS-Net achieves a superior efficiency-accuracy trade-off. By replacing parameter redundancy with efficient frequency-domain feature extraction, our model requires only 5.01 M parameters—approximately 50% of nnU-Net (**Tab. 1**)—without compromising noise resilience. The noise-trained experiment complements the clean-trained noisy-test evaluation by assessing a different aspect of robustness. The clean-trained setting reflects the model's tolerance to unseen noise, whereas the noise-trained setting evaluates whether the model can learn stable representations when low-SNR degradation is already present during training. The consistent performance of FSS-Net under both settings suggests that its robustness is not merely a passive insensitivity to test-time perturbations, but is related to the way frequency-spatial features are organized during learning.

The underlying rationale for this advantage lies in addressing the intrinsic limitations of pure spatial domain processing. In traditional CNNs, speckle noise manifests as high-frequency intensity fluctuations that inherently overlap with vessel boundaries, creating a spatial entanglement that confounds the model. Unlike prior approaches that utilize wavelet denoising as a decoupled preprocessing step—which often incurs the unintended loss of fine edge details—FSS-Net implements a deep architectural integration strategy. By embedding the Wavelet-Enhanced Bottleneck (WEB) and Channel-Spatial-Wavelet Attention (CSWA) modules directly into the feature learning pipeline, the model leverages spectral separability. Since noise is physically isolated in specific high-frequency sub-bands while anatomical structures dominate the low-frequency approximation, this design allows the network to selectively suppress noise artifacts while preserving valid boundary features. This strategy effectively resolves the inherent conflict between denoising and edge preservation that limits traditional spatial methods.

### 5.6. Clinical Relevance and Potential Application Value

Compared with region-overlap metrics such as DSC and IoU, boundary-level accuracy is more directly related to contour-based vascular measurements. For example, lumen area estimation, equivalent diameter calculation, vessel morphology assessment, and wall-related measurements all depend on accurate and continuous boundary localization. In this regard, the addition of HD95 provides a more clinically meaningful evaluation beyond pixel-wise overlap. The proposed method achieved the lowest HD95 of 2.10 ± 0.13 in our experiments, suggesting smaller boundary deviation between the predicted contour and the expert annotation. This indicates that FSS-Net not only improves region-level segmentation performance but also provides more accurate boundary localization, which is important for downstream morphology-related measurements.

Furthermore, the proposed method shows stable performance under both clean and noisy settings, which is important for practical ultrasound applications. In real-world clinical scenarios, image quality may vary across patients, devices, operators, and acquisition conditions. A model that maintains relatively stable segmentation performance under low-SNR conditions may be more suitable for deployment in routine ultrasound analysis workflows. The moderate parameter size of FSS-Net also suggests that it has potential for integration into computer-aided diagnosis systems or ultrasound-assisted quantitative analysis platforms.

Since the dataset only provides annotations for the carotid lumen region and does not include independent plaque masks, measurement errors for plaque area and plaque thickness were not quantitatively analyzed. Future studies will rely on multi-center datasets and standardized expert-validated protocols to further investigate clinically relevant quantitative indicators, including lumen area error, vessel diameter error, wall thickness estimation, and inter-observer variability.

## 6. Conclusion

To resolve the intrinsic conflict between speckle noise suppression and boundary preservation in carotid ultrasound, this study introduces FSS-Net, a framework built upon a Frequency-Spatial Synergy paradigm. By synergizing the Channel-Spatial-Wavelet Attention (CSWA), Laplacian-Guided Adaptive Edge Fusion (LAEF), and Wavelet-Enhanced Bottleneck (WEB) modules, the architecture achieves spectral separability of noise and topological restoration of vessels. This design effectively decouples artifacts from semantic features while enforcing explicit edge constraints. Extensive validation demonstrates that FSS-Net establishes a new benchmark with a Dice Similarity Coefficient of 96.46%. Notably, it exhibits exceptional robustness, maintaining a DSC of 92.19% even under severe noise, while requiring only 5.01 M parameters. This superior efficiency-accuracy trade-off, further corroborated by cross-task validation, confirms FSS-Net as a lightweight, clinically viable solution for precise morphological assessment.

### Declaration of competing interest

The authors declare that there are no conflicts of interest regarding the publication of this paper.

### Acknowledgements

This work was supported in part by the National Natural Science Foundation of China (62466033), in part by the Jiangxi Provincial Natural Science Foundation (20242BAB20070), and in part by the Beijing Natural Science Foundation (8244059).

## References

[1] T. Liang *et al.*, "Evaluation of lipoprotein-associated phospholipase A2, serum amyloid A, and fibrinogen as diagnostic biomarkers for patients with acute cerebral infarction," (in English), *Journal of Clinical Laboratory Analysis,* Article vol. 34, no. 3, p. 7, Mar 2020, Art. no. e23084.

[2] V. L. Feigin *et al.*, "World Stroke Organization (WSO): Global Stroke Fact Sheet 2022 (vol 17, pg 18, 2022)," (in English), *International Journal of Stroke,* Correction vol. 17, no. 4, pp. 478-478, Apr 2022, Art. no. 17474930221080343.

[3] G. A. Roth *et al.*, "Global Burden of Cardiovascular Diseases and Risk Factors, 1990-2019 Update From the GBD 2019 Study," (in English), *Journal of the American College of Cardiology,* Review vol. 76, no. 25, pp. 2982-3021, Dec 22 2020.

[4] X. Li *et al.*, "A deep learning-based system to identify originating mural layer of upper gastrointestinal submucosal tumors under EUS," *ENDOSCOPIC ULTRASOUND,* vol. 12, no. 6, pp. 465-471, NOV-DEC 2023.

[5] J. Iglesias-Garcia *et al.*, "Contrast harmonic endoscopic ultrasound: Instrumentation, echoprocessors, and echoendoscopes," *ENDOSCOPIC ULTRASOUND,* vol. 6, no. 1, pp. 37-42, JAN-FEB 2017.

[6] M. Islam and T. Lee, "An automated extraction of spectral-temporal and spatial-temporal features of EEG for emotion detection," (in eng), no. 2198-4018 (Print).

[7] A. D. Jamthikar *et al.*, " Ensemble Machine Learning and Its Validation for Prediction of Coronary Artery Disease and Acute Coronary Syndrome Using Focused Carotid Ultrasound," *IEEE Transactions on Instrumentation and Measurement,* vol. 71, p. 10, 2022.

[8] S. Li *et al.*, "ERANet: Edge replacement augmentation for semi-supervised meniscus segmentation with prototype consistency alignment and conditional self-training," *Neural Networks,* vol. 196, p. 108337, 2026/04/01/ 2026.

[9] B. Liu *et al.*, "DGSSA: Domain generalization with structural and stylistic augmentation for retinal vessel segmentation," *Neural Networks,* vol. 194, p. 108118, 2026/02/01/ 2026.

[10] Y. L. Zou *et al.*, "MGML: A plug-and-play meta-guided multi-modal learning framework for incomplete multimodal brain tumor segmentation," *NEURAL NETWORKS,* vol. 197, MAY 2026, Art. no. 108525.

[11] Y. Nagaraj *et al.*, "Carotid wall segmentation in longitudinal ultrasound images using structured random forest," (in English), *Computers & Electrical Engineering,* Article vol. 69, pp. 753-767, Jul 2018.

[12] C. X. Zhang *et al.*, "A multimodal artificial intelligence system for the detection and diagnosis of solid pancreatic lesions under EUS," *ENDOSCOPIC ULTRASOUND,* vol. 14, no. 5, pp. 274-281, SEP-OCT 2025.

[13] S. Kiremitci *et al.*, "The role of artificial intelligence and deep learning in determining the histopathological grade of pancreatic neuroendocrine tumors by using EUS images," *ENDOSCOPIC ULTRASOUND,* vol. 14, no. 2, pp. 48-56, MAR-APR 2025.

[14] C. X. Zhang *et al.*, "Risk stratification system of gastrointestinal stromal tumors under EUS elastography based on artificial intelligence," *ENDOSCOPIC ULTRASOUND,* vol. 14, no. 4, pp. 220-227, JUL-AUG 2025.

[15] F. Mao *et al.*, "Segmentation of carotid artery in ultrasound images: Method development and evaluation technique," (in English), *Medical Physics,* Article vol. 27, no. 8, pp. 1961-1970, Aug 2000.

[16] S. Golemati *et al.*, "Using the hough transform to segment ultrasound images of longitudinal and transverse sections of the carotid artery," (in English), *Ultrasound in Medicine and Biology,* Article vol. 33, no. 12, pp. 1918-1932, Dec 2007.

[17] C. Azzopardi *et al.*, "Automatic carotid ultrasound segmentation using deep convolutional neural networks and phase congruency maps," in *IEEE 14th International Symposium on Biomedical Imaging (ISBI) - From Nano to Macro*, Melbourne, AUSTRALIA, 2017, pp. 624-628, NEW YORK: Ieee, 2017.

[18] P. K. Jain *et al.*, "Localization of common carotid artery transverse section in B-mode ultrasound images using faster RCNN: a deep learning approach," (in English), *Medical & Biological Engineering & Computing,* Article vol. 58, no. 3, pp. 471-482, Mar 2020.

[19] N. Ottakath *et al.*, "Bi-attention DoubleUNet: A deep learning approach for carotid artery segmentation in transverse view images for non-invasive stenosis diagnosis," *Biomedical Signal Processing and Control,* vol. 94, 2024.

[20] J. de Ruijter *et al.*, "A Generalized Approach for Automatic 3-D Geometry Assessment of Blood Vessels in Transverse Ultrasound Images Using Convolutional Neural Networks," *IEEE Transactions on Ultrasonics, Ferroelectrics, and Frequency Control,* vol. 68, no. 11, pp. 3326-3335, 2021.

[21] K. Wang *et al.*, "EANet: Iterative edge attention network for medical image segmentation," *Pattern Recognition,* vol. 127, 2022.

[22] J. Liu *et al.*, "Toward automated right ventricle segmentation via edge feature-induced self-attention multiscale feature aggregation full convolution network," *IEEE Transactions on Instrumentation and Measurement,* vol. 72, no. -, p. 12, 2023.

[23] C. C. Yeung and K. M. Lam, " Attentive Boundary-Aware Fusion for Defect Semantic Segmentation Using Transformer," *IEEE Transactions on Instrumentation and Measurement,* vol. 72, p. 13, 2023.

[24] M. Sun *et al.*, "Attention Mechanism Enhanced Multi-layer Edge Perception Network for Deep Semantic Medical Segmentation," *Cognitive Computation,* vol. 15, no. 1, pp. 348-358, 2023.

[25] S. W. J. Park, J.-Y. Lee, and I. S. Kweon, "BAM: Bottleneck Attention Module," in *British Machine Vision Conference (BMVC)*, 2018.

[26] S. H. Woo *et al.*, "CBAM: Convolutional Block Attention Module," in *15th European Conference on Computer Vision (ECCV)*, Munich, GERMANY, 2018, vol. 11211, pp. 3-19, CHAM: Springer International Publishing Ag, 2018.

[27] X. Y. Zhao *et al.*, "Wavelet-Attention CNN for image classification," (in English), *Multimedia Systems,* Article vol. 28, no. 3, pp. 915-924, Jun 2022.

[28] J. Cheng *et al.*, "WaveNet-SF: A hybrid network for retinal disease detection based on wavelet transform in spatial-frequency domain," *Neural Netw,* vol. 194, p. 108189, Oct 3 2025.

[29] Z. Qi *et al.*, "MSLI-Net: retinal disease detection network based on multi-segment localization and multi-scale interaction," *Front Cell Dev Biol,* vol. 13, p. 1608325, 2025.

[30] C. Tian *et al.*, "Multi-stage image denoising with the wavelet transform," *Pattern Recognition,* vol. 134, 2023.

[31] G. Xu *et al.*, "Haar wavelet downsampling: A simple but effective downsampling module for semantic segmentation," *Pattern Recognition,* vol. 143, 2023.

[32] T. Yao *et al.*, "Wave-ViT: Unifying Wavelet and Transformers for Visual Representation Learning," vol. abs/2207.04978, 2022.

[33] G. Xu *et al.*, "Haar wavelet downsampling: A simple but effective downsampling module for semantic segmentation," *Pattern Recognition,* vol. 143, p. 109819, 2023/11/01/ 2023.

[34] Y. Zhao *et al.*, "WRANet: wavelet integrated residual attention U-Net network for medical image segmentation," *Complex & Intelligent Systems,* vol. 9, no. 6, pp. 6971-6983, 2023/12/01 2023.

[35] Y. Bi *et al.*, "MI-SegNet: Mutual Information-Based US Segmentation for Unseen Domain Generalization," in *Medical*

*Image Computing and Computer Assisted Intervention – MICCAI 2023*, Cham, 2023, pp. 130-140: Springer Nature Switzerland.

[36] O. Ronneberger *et al.*, "U-Net: Convolutional Networks for Biomedical Image Segmentation," in *18th International Conference on Medical Image Computing and Computer-Assisted Intervention (MICCAI)*, Munich, GERMANY, 2015, vol. 9351, pp. 234-241, CHAM: Springer International Publishing Ag, 2015.

[37] X. Li *et al.*, "Selective Kernel Networks," presented at the 2019 IEEE/CVF Conference on Computer Vision and Pattern Recognition (CVPR), 2019.

[38] S. Zhang *et al.*, "Selective kernel convolution deep residual network based on channel-spatial attention mechanism and feature fusion for mechanical fault diagnosis," *ISA Transactions,* vol. 133, pp. 369-383, 2023.

[39] A. Momot, "Common Carotid Artery Ultrasound Images," M. Data, Ed., ed, 2022.

[40] W. G. Al-Dhabyani, Mohammed Khaled, H. Fahmy, Aly, "Dataset of breast ultrasound images," *Data in Brief,* vol. 28, p. 104863, 2020.

[41] L. C. E. Chen *et al.*, "Encoder-Decoder with Atrous Separable Convolution for Semantic Image Segmentation," in *15th European Conference on Computer Vision (ECCV)*, Munich, GERMANY, 2018, vol. 11211, pp. 833-851, CHAM: Springer International Publishing Ag, 2018.

[42] J. N. Chen *et al.*, "TransUNet: Rethinking the U-Net architecture design for medical image segmentation through the lens of transformers," (in English), *Medical Image Analysis,* Article vol. 97, p. 10, Oct 2024, Art. no. 103280.

[43] N. Ibtehaz and M. S. Rahman, "MultiResUNet : Rethinking the U-Net architecture for multimodal biomedical image segmentation," *Neural Netw,* vol. 121, pp. 74-87, Jan 2020.

[44] S. K. Wazir, D., "Rethinking Decoder Design: Improving Biomarker Segmentation Using Depth-to-Space Restoration and Residual Linear Attention," presented at the Proceedings of the IEEE/CVF Conference on Computer Vision and Pattern Recognition (CVPR) 2025, 2025. Available: https://github.com/saadwazir/MCADS-Decoder

[45] Z. Jia *et al.*, "CondSeg: Ellipse Estimation of Pupil and Iris via Conditioned Segmentation," in *18th European Conference on Computer Vision (ECCV)*, Milan, ITALY, 2024, vol. 15645, pp. 185-198, 2025.

[46] C. Jiang *et al.*, "MHAHF-UNet: a multi-scale hybrid attention hierarchy fusion network for carotid artery segmentation," *Int J Comput Assist Radiol Surg,* Jun 17 2025.

[47] Y. Li *et al.*, "FRDD-Net: Automated carotid plaque ultrasound images segmentation using feature remapping and dense decoding," *Sensors (Basel),* vol. 22, no. 3, Jan 24 2022.

[48] F. Isensee *et al.*, "nnU-Net: a self-configuring method for deep learning-based biomedical image segmentation," *Nature Methods,* vol. 18, no. 2, p. 211, 2021.

[49] O. S. Oktay, J.; Le Folgoc, L.; Lee, M.; Heinrich, M.; Misawa, K.; Mori, K.; McDonagh, S.; Hammerla, N. Y.; Kainz, B.; Glocker, B.; Rueckert, D., "Attention U-Net: Learning Where to Look for the Pancreas," *arXiv preprint arXiv:1804.03999,* 2018.

[50] Z. X. Zhang *et al.*, "Road Extraction by Deep Residual U-Net," (in English), *Ieee Geoscience and Remote Sensing Letters,* Article vol. 15, no. 5, pp. 749-753, May 2018.

[51] Z. Q. Ma *et al.*, "HCTNet: A Hybrid ConvNet-Transformer Network for Retinal Optical Coherence Tomography Image Classification," (in English), *Biosensors-Basel,* Article vol. 12, no. 7, p. 15, Jul 2022, Art. no. 542.

[52] A. Roy *et al.*, "EU2-Net: A Parameter Efficient Ensemble Model With Attention-Aided Triple Feature Fusion for Tumor Segmentation in Breast Ultrasound Images," *IEEE Transactions on Instrumentation and Measurement,* vol. 73, no. -, p. 7, 2024.